\documentclass[conference]{IEEEtran}
\IEEEoverridecommandlockouts
\usepackage{cite}
\usepackage{amsmath,amssymb,amsfonts}
\usepackage{algorithmic}
\usepackage{graphicx}
\usepackage{textcomp}
\usepackage{xcolor}
\usepackage{url}
\def\BibTeX{{\rm B\kern-.05em{\sc i\kern-.025em b}\kern-.08em
    T\kern-.1667em\lower.7ex\hbox{E}\kern-.125emX}}
\begin{document}

\title{Policy Gradient Stock GAN for Realistic Discrete Order Data Generation in Financial Markets}

\author{\IEEEauthorblockN{Masanori HIRANO}
\IEEEauthorblockA{\textit{School of Engineering} \\
\textit{The University of Tokyo}\\
Tokyo, Japan\\
research@mhirano.jp}
\and
\IEEEauthorblockN{Hiroki SAKAJI}
\IEEEauthorblockA{\textit{School of Engineering} \\
\textit{The University of Tokyo}\\
Tokyo, Japan\\
sakaji@sys.t.u-tokyo.ac.jp}
\and
\IEEEauthorblockN{Kiyoshi IZUMI}
\IEEEauthorblockA{\textit{School of Engineering} \\
\textit{The University of Tokyo}\\
Tokyo, Japan\\
izumi@sys.t.u-tokyo.ac.jp}
}

\maketitle

\begin{abstract}
	This study proposes a new generative adversarial network (GAN) for generating realistic orders in financial markets.
	In some previous works, GANs for financial markets generated fake orders in continuous spaces because of GAN architectures' learning limitations.
	However, in reality, the orders are discrete, such as order prices, which has minimum order price unit, or order types.
	Thus, we change the generation method to place the generated fake orders into discrete spaces in this study.
	Because this change disabled the ordinary GAN learning algorithm, this study employed a policy gradient, frequently used in reinforcement learning, for the learning algorithm.
	Through our experiments, we show that our proposed model outperforms previous models in generated order distribution.
	As an additional benefit of introducing the policy gradient, the entropy of the generated policy can be used to check GAN's learning status.
	In the future, higher performance GANs, better evaluation methods, or the applications of our GANs can be addressed.
\end{abstract}


\section{Introduction}
In financial markets, the realized order time series is a very tiny part of possible state spaces.
In other words, there is only one path that has been realized among the various paths of market movements.
Although the pattern of one order is limited, the state space could be huge when the orders are compiled.
In addition, due to the lack of stationarity of financial markets, all possible states could not be realized in the past data, or data for some possible states could not be obtained enough.

Obtaining much more data would give us various advantages.
For example, we could improve trading strategy performance based on much more data.
Backtesting with more data could improve the more accurate estimation of risk levels.
Moreover, a more accurate portfolio evaluation using various augmented time-series price paths would be available.

Many studies have been conducted to address the insufficiency of data.
% To this end, the prominent approach is data augmentation via GANs.
% Undo
To this end, the prominent approaches are artificial market simulations and data augmentation via GANs.

Artificial market simulations aim to simulate virtual markets under hypothetical situations and examine those situations that have not occurred in the actual financial market.
Events such as financial crises are rare, but varied. However, data about such events are insufficient.
Additionally, the effects of external factors, such as new regulations, remain unknown.
Therefore, by controlling the situations in artificial market simulations, we receive insights and benefits.

The other approach using GANs aims to make realistic order time series to augment past data.
This approach is more pragmatic than artificial market simulations.
For predictions, deep learning approaches are gaining popularity; however, they require a plenty of data.
GANs can fill this need.
% end of undo

This study focuses on 
%undo
the latter approach, that is,
%end of undo
GANs for making realistic order time series in financial markets, especially in stock markets.

Although some research focuses on GANs for stock markets, the generated fake data remain unrealistic.
Stock-GAN (S-GAN) \cite{Li2020} and Market GAN \cite{Naritomi2020} are some previous works for stock markets.
To train GANs, gradient connection between the generator and the critic is required.
This requirement causes the generator outputs (the generated fake data) to be unrealistic, thereby affecting the learnability.
In the case of S-GAN, the values generated for buy/sell, types, prices, and volumes of orders, are continuous.
It is easy for humans to identify such fake data because the actual orders are not continuous.
While Market GAN generates discrete values and calculates the class probabilities, the generated fake data remain unrealistic because humans can easily distinguish between probabilistic class data and real data.
The most important problem of current GANs for financial markets is that the generated fake data are placed in continuous space.
For example, you can notice that the following generated orders are fake:
\begin{itemize}
	\item An order is buy orders in the probability of 0.6. (Occured in S-GAN and Market GAN)
	\item An order whose price is \$100.0123 in the market, whose minimum order price unit (price tick size) is \$0.01. (Occured in S-GAN)
\end{itemize}
In our opinion, also generated orders should be order-system-acceptable orders.
Please imagine when you are trying to make a new order in a real market.
It is possible that you make a strategy to make a new order with probability in your mind.
However, when you submit your order via the ordering system, you have to decide your order is buy or sell.

In addition, there is also a major problem with treating buy/sell and order types as continuous values, and joining their state spaces next to each other.
If buy/sell was expressed as $[0, 1]$, buy and sell would be continuously joined in terms of their state space.
For example, suppose the order is 100 shares at \$200. In that case, when the best quote is \$190, the meaning is completely different between sell and buy; the buy at \$200 is effectively a take order; the sell at \$200 is effectively a make order.

Thus, this discreteness of financial market orders should be incorporated into the design of the GANs, especially in the generators.

The most important part of solving the problem is the necessity of gradient connection between the generator and critic of GANs for training.
Suppose the gradient connection between the generator and critic is not required.
In that case, the generator can make more realistic orders without continuous unrealistic values or make sampled fake orders from class probability estimation.

For solving the current issue, this study proposes Policy Gradient Stock GAN (PGSGAN), a new GAN learning method for stock markets using policy gradient.
Policy gradient is a learning algorithm that is frequently used in reinforcement learning.
By incorporating the relationship between the generator and critic of GANs into the concept of reinforcement learning, we make policy gradient available in GANs for stock markets.
This introduction enables GANs to remove the gradient connectivity between their generators and critics, and make more realistic discrete orders.

Consequently, we successfully design the generator output of the GAN with more realistic market trading rules and improve the generation performance.
Because this technology enables to augment of more realistic data, it is expected to improve the learnability of prediction tasks and other tasks via machine learning methods.

Although PGSGAN is designed according to the rules of the Tokyo Stock Exchange (TSE), it can also be applied to other markets with some small changes.

\section{Related Work}
% undo
As mentioned above, there are two main types of data augmentation approaches in financial markets.

In the artificial market simulation context, Maeda {\it et al.} \cite{maeda2020deep} attempted to make a model learning the better trading strategy via augmented data by artificial market.
Returning to the basics of artificial market simulation, Edmonds {\it et al.} \cite{Edmonds2005} argued that agent-based simulation is useful for social sciences.
The importance of agent-based simulation, especially for the financial markets, was discussed in \cite{Farmer2009,Battiston2016}.
Muzuta \cite{Mizuta2019} demonstrated that a multi-agent simulation for the financial market could contribute to the implementation of rules and regulations in actual financial markets.
Indeed, Mizuta {\it et al.} \cite{Mizuta2016} tested the effect of price tick size, that is, the price unit for orders, which led to a discussion of tick size devaluation in the Tokyo Stock Exchange Market, based on the data from an artificial market simulation.
Hirano {\it et al.} \cite{Hirano2020c} assessed the effect of the regulation of the capital adequacy ratio (CAR) and observed the risk of market price shock and depression due to CAR regulation from the generated data under the hypothetical situation realized in their artificial market simulation.
Although artificial market simulations do not seem to augment data, these can be called thus, because they generate data and use it for discussion.

The other approach employed in this study is GANs.
% end of undo
% As mentioned above,
S-GAN proposed in \cite{Li2020} was a GAN for stock markets, designed to generate realistic order time series and also generate best prices without real order book data by continuous double auction (CDA) network.
Naritomi {\it et al.} \cite{Naritomi2020} showed a basic GAN model for stock markets and that its generated data are beneficial for predicting the future price movement.
Moreover, there exist studies which attempted to use GAN architecture directly for future price predictions \cite{Zhou2018,Zhang2019}.
As observed from those studies, currently, GANs' approaches are more practical.

The technology related to GANs has been improving, in the following respects.
Initially, Goodfellow {\it et al.} \cite{Goodfellow2014} proposed the original GAN.
Then, Mirza {\it et al.} \cite{Mirza2014} proposed the conditional GAN, whose idea was also used in this study, using conditional inputs.
Radford {\it et al.} \cite{Radford2015} proposed a deep convolutional GAN (DCGAN), which was also used in this study as a comparison model.
As other learning architectures, the least-squares GAN \cite{mao2017least}, generalized f-GAN \cite{Nowozin2016}, Laplacian pyramid GAN \cite{Denton2015}, variational autoencoder GAN \cite{Boesen2016}, image-to-image translation GAN (known as pix2pix) \cite{Isola2017}, self-attention GAN \cite{Zhang2019}, cycle GAN \cite{Chu2017} for image translations, style GAN \cite{Karras2019} for style converts, and progressive growing GAN \cite{Karras2018} for high-resolution images were proposed.
Moreover, Yu {\it et al.}\cite{Yu2017} proposed SeqGAN for sequence generation, such as text or music, using policy gradient.
Although it could seems smilar to our study, our GAN is not generating sequences.
As an extension of GANs, adversarial feature learning \cite{Donahue2016} for embedding images into vectors and anomaly detections based on GAN \cite{Schlegl2017,Zenati2018,Deecke2018} were proposed.
Wasserstein GAN (WGAN) \cite{Arjovsky2017} was suggested based on the discussion of the learning stability of GANs \cite{Arjovsky2017-theory}.
Our study is based on WGAN.
Gradient penalty \cite{Gulrajani2017} and spectral normalization \cite{Miyato2018} were the tools proposed for stabilizing the WGAN.
In this study, thus, we employed spectral normalization.

\section{Models}
\subsection{Policy Gradient Stock GAN (PGSGAN)}
PGSGAN uses historical data and current conditions to generate the next order in financial markets.
It is based on GAN \cite{Goodfellow2014} and policy gradient theorem \cite{witten1977adaptive,barto1983neuronlike,sutton2000policy}.
More technically, we used Wasserstein GAN (WGAN) \cite{Arjovsky2017} as a basis, REINFORCE \cite{williams1992simple} with a baseline as a policy gradient algorithm, and convolutional neural network (CNN) as a part of our neural network.
We also used batch normalization \cite{Ioffe2015}, layer normalization \cite{Ba2016}, and spectral normalization \cite{Miyato2018} for stabilizing the learning.

\subsubsection{\bf PGSGAN Architecture}
\begin{figure}[htb]
	\centering
	\vspace{-3mm}
	\includegraphics[width=\columnwidth]{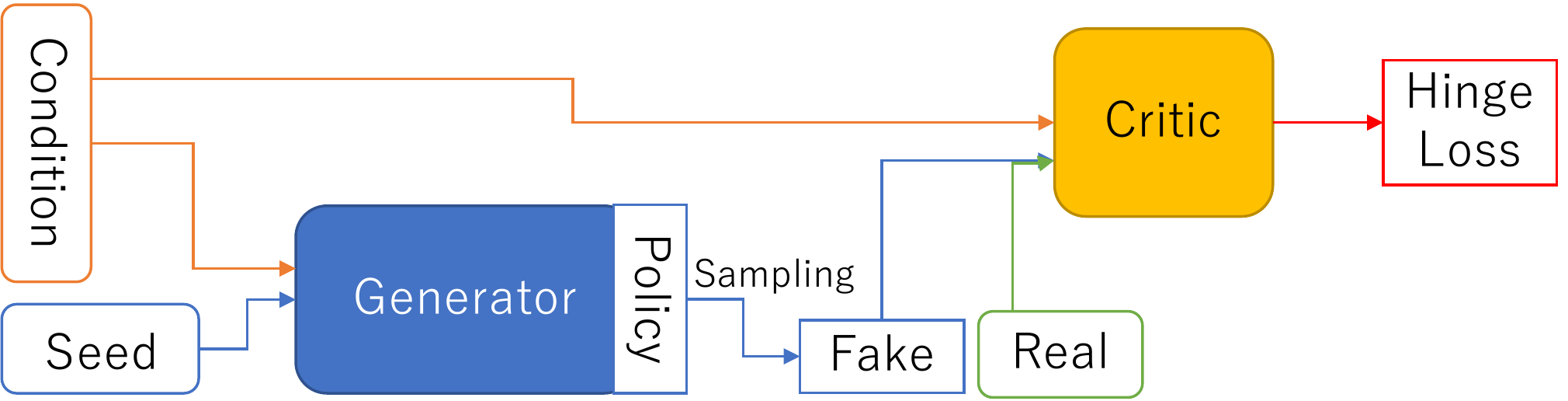}
	\vspace{-6mm}
	\caption{The outline of PGSGAN}
	\vspace{-2mm}
	\label{fig:PGSGAN}
\end{figure}

Figure \ref{fig:PGSGAN} indicates the outline of our PGSGAN.

The generators accept conditional data (historical data of markets) and random seeds for a generation.
In our experiments, the conditional data comprises the last 20 order time series and the current best sell and buy prices.
The order time series contains information, such as buy/sell, new/cancel, market order (or not), price (ticks from the best price), volume (scaled with dividing by the minimum volume unit), and best prices before the order.
The fake seed has 128 dimensions and is randomly generated.

Then, the generator makes a policy for generating the fake next order.
In our experiment, the generator used 14 convolutional layers and 5 linear layers to make the following policy:% (details presented in the supplementary document):
\newcounter{footnotecommon1}
\begin{itemize}
	\item Sell or Buy -- 2 classes (probability)
	\item New or Cancel -- 2 classes (probability)
	\item is a market order? (Is MO) -- 2 classes (probability)
	\item relative price (ticks from the best price) -- 40 classes (probabilities for 0 -- 39) \footnote{When the value is equal or more than 40, it was regarded as 40th class. Although these cases (40 ticks over or more than 4000 shares) could happen, the percentages are very limited in TSE. Moreover, these cases usually occurred by events outside markets themselves; thus, we ignore the detailed modeling of these cases in this study.}\setcounter{footnotecommon1}{\value{footnote}}
	\item volume (scaled by dividing by minimum volume unit) -- 40 classes (probabilities for 0 -- 39)\footnotemark[\value{footnotecommon1}]
\end{itemize}

After calculating the probabilistic policy, the generator makes a fake next order by weighted sampling, according to the policy.

However, the critic only maps the inputs into the scalar under the 1-Lipschitz constraint.
This is the same as the basic WGAN.
The critic accepts two inputs: conditional data (the historical data, same as the generator), and either the generated fake next order or the real next order.
In our experiments, the real next orders were also converted into the above-explained range.
This implies that only the price and volume of the real data were mapped into 0-39 (integer).
If the real order is a market order, the price value is set to 0.

\subsubsection{\bf PGSGAN Learning Mechanism}
Because of the sampling process, the gradient connection between the generator and the critic is lost.
Traditional GANs, especially all GANs for stock markets, rely on the gradient connection between their generators and critic for training their generator.
However, our PGSGAN completely abandoned the connection and disabled the traditional learning theory for the generator, because of the sampling process for generating a fake order, based on a generated policy.

Thus, as a new learning theory for the generator, we employed the policy gradient widely used in reinforcement learning.

In the following, we use the notations:
\begin{itemize}
	\item $z$: random variables (seed for generator. In our experiment, $z \in \mathbb{R}^{128}$.)
	\item $\mathbb{P}_z$: the distribution of random variables
	\item $\mathbb{P}_r$: the distribution of real data
	\item $C(x)$: the critic as a function. The output is scalar. Here, $x$ is a given input from the outside.\footnote{Correctly, it also accepts conditional data, but it is ignored in this notation for simplicity.}\setcounter{footnotecommon1}{\value{footnote}}
	\item $G(z)$: the generator as a function. The output is a policy. Usually, the generator accepts random seeds.\footnotemark[\value{footnotecommon1}]
	\item $\tilde{x} \sim G(z)$: the sampled fake order $\tilde{x}$ follows the policy generated by $G(z)$.
	\item $\theta_C, \theta_G$: params in the critic and generator, respectively.
	\item $L_C, L_G$: loss function for the critic and generator.
	\item $||f||_{L\leq 1}$: 1-Lipschitz constraint for any function $f$.
	\item $p_{G(z)}(\tilde{x})$: probability for the sampled fake order $\tilde{x}$ according to the generated policy $G(z)$.
	\item $\mathrm{NLL}_{G(z)}(\tilde{x})$: negative log-likelihood for the sampled fake order $\tilde{x}$ according to the generated policy $G(z)$. $\mathrm{NLL}_{G(z)}(\tilde{x}) = - \ln{\{p_{G(z)}(\tilde{x})\}}$.
\end{itemize}

At first, PGSGAN will solve the following minimax game:
\begin{eqnarray}
	\min_G \max_{||C||_{L\leq 1}} L_{\mathrm{GAN}}(G, C),
\end{eqnarray}
where
\begin{eqnarray}
	L_{\mathrm{GAN}}(G, C) := \mathbb{E}_{x\sim \mathbb{P}_r}\left[C(x)\right] - \mathbb{E}_{z\sim \mathbb{P}_z}\left[\mathbb{E}_{\tilde{x}\sim G(z)}\left[C(\tilde{x})\right]\right].\nonumber
\end{eqnarray}
This form is similar to the original form of WGAN.
However, the sampling term $\mathbb{E}_{\tilde{x}\sim G(z)}$ has been added.
This change is substantial for generator learning.

For the critic, the objective function is
\begin{eqnarray}
	\max_{||C||_{L\leq 1}} \left\{\mathbb{E}_{x\sim \mathbb{P}_r}\left[C(x)\right] - \mathbb{E}_{\tilde{x}\sim G(z)}\left[C(\tilde{x})\right]\right\}
\end{eqnarray}
because the generator and its seeds do not matter to the critic.
Thus, the loss function for the critic is:
\begin{eqnarray}
	L_C := \mathbb{E}_{\tilde{x}\sim G(z)}\left[C(\tilde{x})\right]- \mathbb{E}_{x\sim \mathbb{P}_r}\left[C(x)\right].
\end{eqnarray}
These are the same as WGAN because the generator does not matter for the critic; only the fake data affect the critic.

\begin{figure}[htb]
	\centering
	\includegraphics[width=\columnwidth]{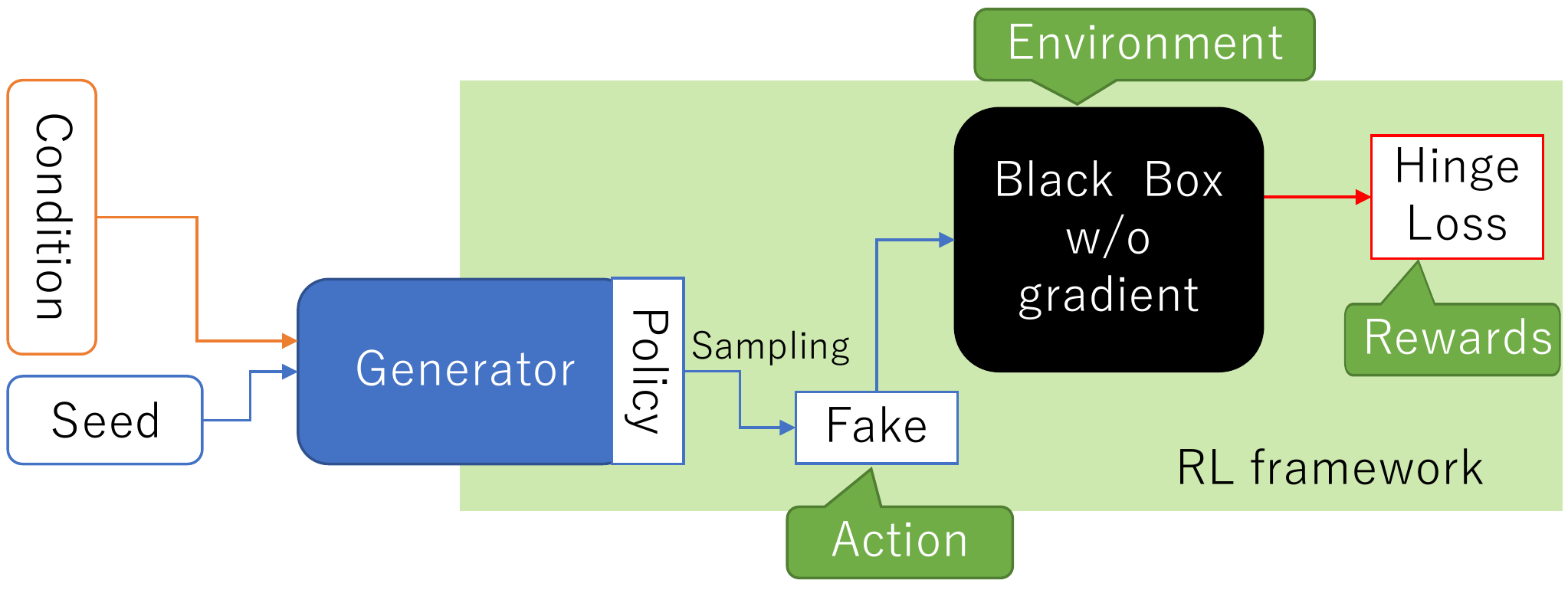}
	\vspace{-6mm}
	\caption{Outline of a generator's learning in terms of a reinforcement learning (RL) framework}
	\label{fig:PGSGAN-RL}
\end{figure}

In contrast, the learning theory for the generator is complicated.
The outline of the learning is shown in figure \ref{fig:PGSGAN-RL}.

The generator's objective function is:
\begin{eqnarray}
	\min_{G} \left\{\mathbb{E}_{x\sim \mathbb{P}_r}\left[C(x)\right] - \mathbb{E}_{z\sim \mathbb{P}_z}\left[\mathbb{E}_{\tilde{x}\sim G(z)}\left[C(\tilde{x})\right]\right]\right\}.
\end{eqnarray}
Like the original WGAN, the first term of this equation is unchangeable for the generator.
Thus, the objective function is re-written as:
\begin{eqnarray}
	\max_{G} \mathbb{E}_{z\sim \mathbb{P}_z}\left[\mathbb{E}_{\tilde{x}\sim G(z)}\left[C(\tilde{x})\right]\right].
\end{eqnarray}
This objective function cannot be converted into a backpropagation of a neural network because of the lack of gradient connection for the generator.
Thus, here, we employ REINFORCE, one of the policy gradient methods, as a learning algorithm from reinforcement learning.

As figure \ref{fig:PGSGAN-RL} shows, the form of the generator can be thought of as reinforcement learning, in which the generator is an actor pursuing higher rewards, generating a policy for action.
Then, according to the generated policy, an action is taken: making fake orders.
Through the unknown environment, then, the action makes a reward: the output from the critic.
Finally, according to the rewards, the actor, that is, the generator, is updated.

According to REINFORCE, the parameter is updated as:
\begin{eqnarray}
	\theta \leftarrow \theta + \alpha \cdot G \nabla_\theta \ln \pi_\theta (a | s)\label{eq:RAINFORCE},
\end{eqnarray}
where $\theta$ is model parameter, $\alpha$ is the learning rate, $G$ is the return (usually the sum of discounted future rewards; however, in this study, just upcoming reward itself caused by action $a$), $\pi_\theta (a|s)$ is a probability of action $a$ under the state $s$ according to the current policy $\pi_\theta$.
By introducing baseline, equation \ref{eq:RAINFORCE} is changed to
\begin{eqnarray}
	\theta \leftarrow \theta + \alpha \cdot (G - B) \nabla_\theta \ln \pi_\theta (a | s)\label{eq:RAINFORCE-baseline},
\end{eqnarray}
where $B$ is the baseline (in this study, we employ mean of $G$ in one batch.)

By applying REINFORCE for PGSGAN, the parameter update of the generator is:
\begin{eqnarray}
	\theta_G \leftarrow \theta_G + \alpha (C(\tilde{x}) - B) \nabla_{\theta_G} \ln p_{G(z)}(\tilde{x}),
\end{eqnarray}
where $\tilde{x} \sim G(z)$, $z \sim \mathbb{P}_z$, and $B$ is the mean of $C(\tilde{x})$ among a learning batch.
Thus, the loss function for the generator is defined as:
\begin{eqnarray}
	L_G &:=& - (C(\tilde{x}) - B) \ln p_{G(z)}(\tilde{x})\\
	&=& (C(\tilde{x}) - B) \cdot \mathrm{NLL}_{G(z)} (\tilde{x})\label{eq:loss_g}.
\end{eqnarray}

Therefore, the generator is enabled to learn.

\subsubsection{\bf Additional Note and Actual Implementation for PGSGAN}
In PGSGAN, we employ spectral normalization in all layers, which is required to realize 1-Lipschitz constraint in critics.
However, for learning stability, we use it also in the generator.

The layers processing the conditional data in both the generator and the critic have the same architecture; however, they are trained separately and not shared.

\begin{figure*}[htbp]
	\centering
	\includegraphics[width=0.95\linewidth]{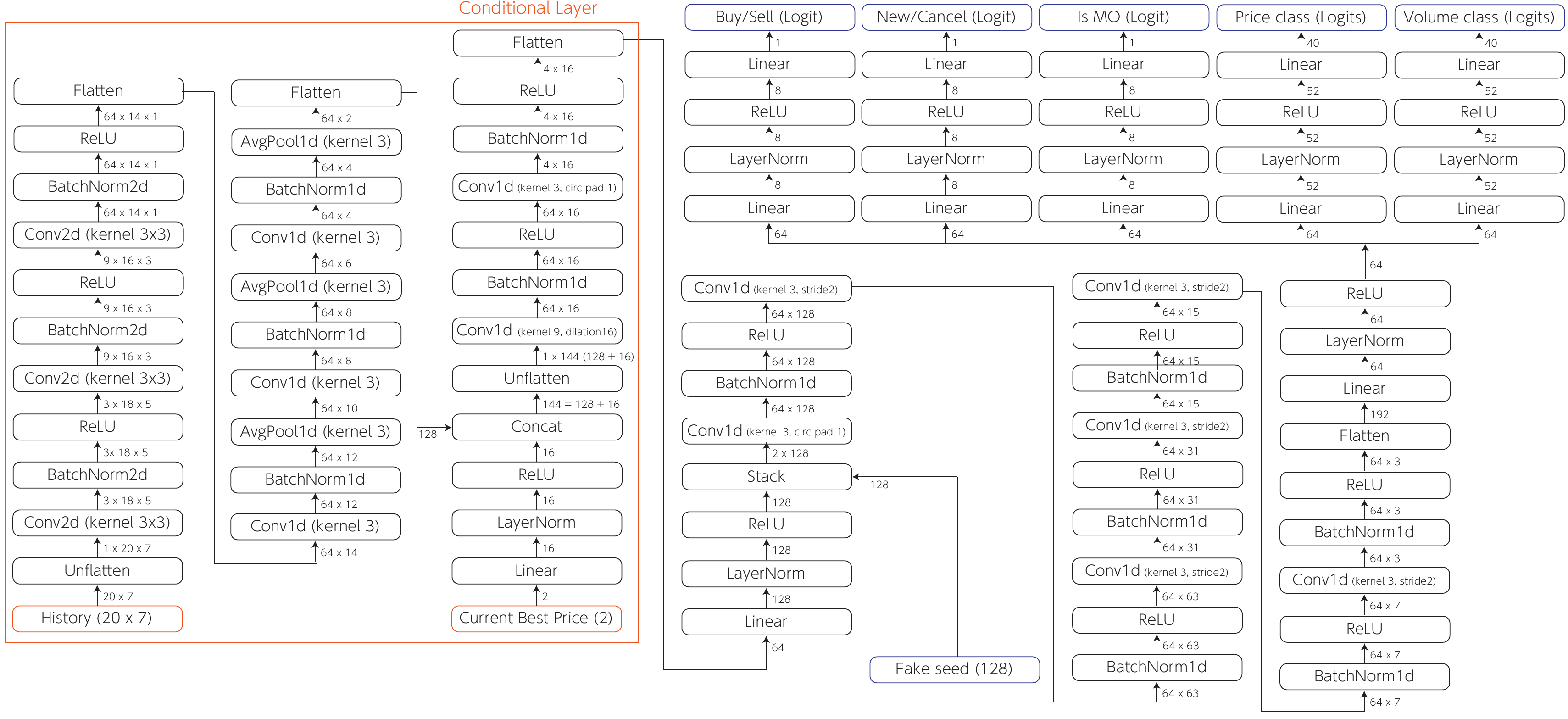}
	\vspace{-4mm}
	\caption{The detailed architectures of the generator}
	\label{fig:generator}
\end{figure*}

\begin{figure*}[htbp]
	\centering
	\includegraphics[width=0.95\linewidth]{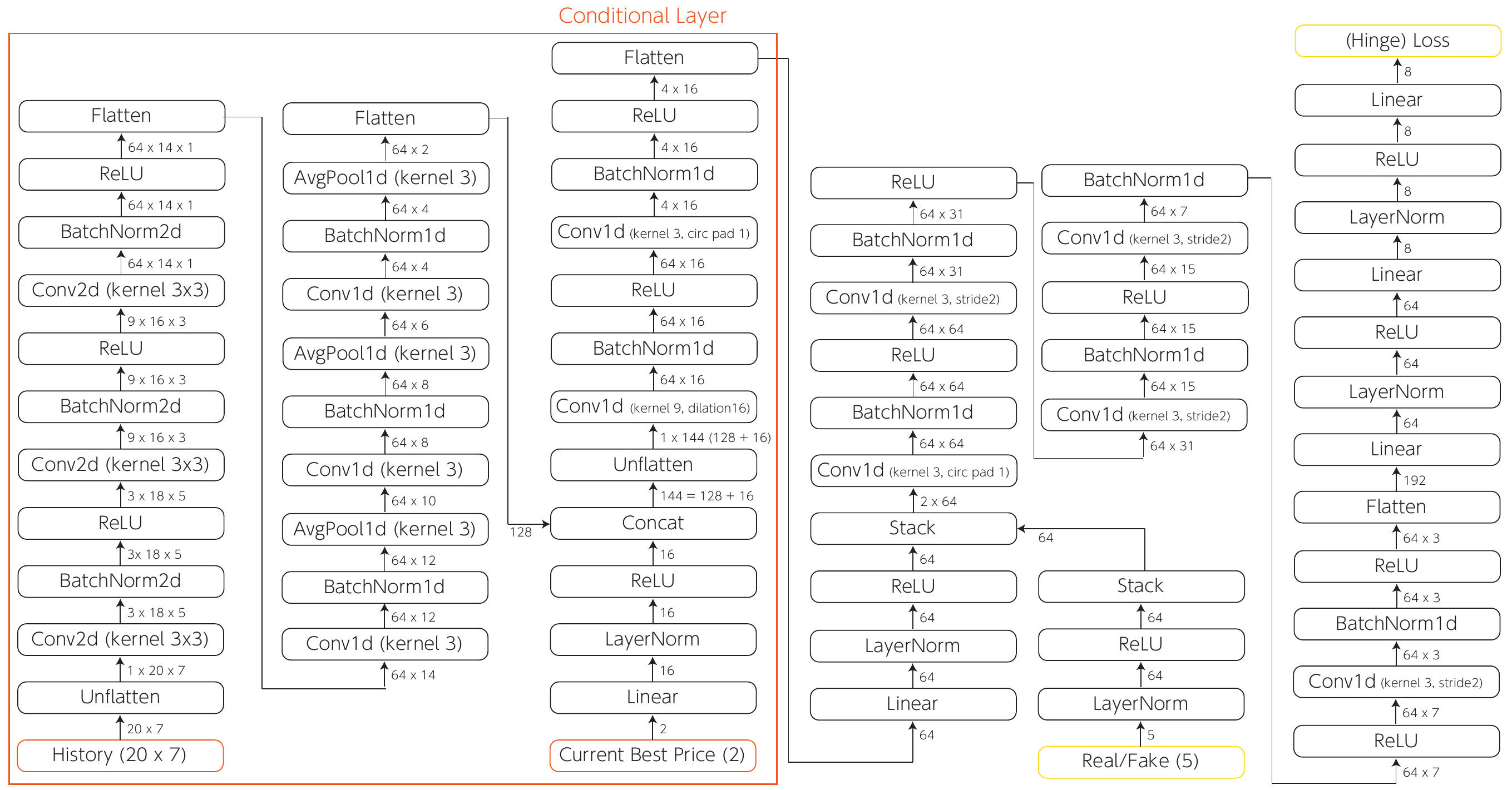}
	\vspace{-3mm}
	\caption{The detailed architectures of the critic}
	\label{fig:critic}
\end{figure*}

In our implementation (including the common architectures for conditional data, which has 44,750 parameters), the number of parameters in the critic and generator are 113,071 and 141,625, respectively.
The detailed implementations are shown in figures \ref{fig:generator} and \ref{fig:critic}.

Figure \ref{fig:generator} shows the details of our generator.
We employ CNN as a basic foundation of our model.
In the conditional layer, for the latter processing part of historical data, we employ average pooling.
It is because this part processes data in the direction of time sequence.
In very high-frequency trading, the sequence of some orders is not significant.
Thus, we employ average pooling to buffer these orders.
After the concatenate of processed historical data and the current best price, we employ a dilated convolution \cite{yu2015multi} and circled convolution, because the concatenated two inputs should be mixed equivalently.
As the outputs, we employ logits for the convenience of calculation.
% As shown in the main paper, the loss function for the generator is calculated as
% \begin{eqnarray}
% 	L_G &:=& - (C(\tilde{x}) - B) \ln P_{G(z)}(\tilde{x})\\
% 	&=& (C(\tilde{x}) - B) \cdot \mathrm{NLL}_{G(z)} (\tilde{x}).
% \end{eqnarray}
As mentioned above, the loss function for the generator is calculated as equation \ref{eq:loss_g}. 
Thus, for compatibility of the negative log-likelihood (NLL), the logits are best for less computational error.
Thus, for making actual policies, these logits are put into sigmoid or softmax.
%As mentioned in the main paper, 
As mentioned above,
we also set sell/buy, new/cancel, and whether the order is market order (MO), as two classes of output.
However, for the convenience of calculation, the outputs have one class.
Thus, for making the policy, we convert them into two classes.
Moreover, in all layers, spectral normalization \cite{Miyato2018} is applied.

Contrary, figure \ref{fig:critic} shows the details of the critic.
Basic architectures are almost the same as the generator, except for the final layers and each dimension.

\subsection{Policy Gradient Stock GAN with Hinge Loss (PGSGAN-HL)}
We also implement the PGSGAN with Hinge loss.
Originally, Hinge loss was used in WGAN in Geometric GAN \cite{Lim2017}.
Hinge loss is defined as
\begin{eqnarray}
	\left\{
	\begin{array}{ll}
		\max (x + 1, 0) & \mathrm{(critic~training~for~fake~data)} \\
		\max (1 - x, 0) & \mathrm{(critic~training~for~real~data)} \\
		x               & \mathrm{(generator~training)}
	\end{array}
	\right.
\end{eqnarray}
and insert into the last of the critic layers.

The others are the same as PGSGAN.

\subsection{Comparative Models}
\subsubsection{\bf Stock GAN (S-GAN)}
Stock GAN (S-GAN) was proposed by \cite{Li2020}, and based on WGAN-GP \cite{Gulrajani2017}.
S-GAN has LSTM for processing conditional data (historical data) and CNN for processing the LSTM output, and either seed for a generation or the next order (fake/real) for a critic.
In this study, we replicate this model as a comparative one.
However, to evaluate fairly, we modify some of its architecture:
\begin{itemize}
	\item Deletion of continuous double auction (CDA) network: This network was originally employed for updating best prices after the new order. However, in this study, we assume a situation that can use all market data. Thus, the best price estimation by CDA network is not required.
	\item Deletion of time signal: The original S-GAN accepts the time signal, which aims to identify when, in one day, the order was placed among the divided 24 classes as one of the inputs. However, the TSE, which we target in this study, has only 2.5-hours sessions (2 sessions per day) and does not have 24/7 markets. Thus, we decide to delete this.
	\item Price processed in this study changed from absolute value to relative scaled price: In the original study, it was generated for a very limited period of time (assumed to be about one day); thus, the importance of relativizing the price level was not too high. However, in this study, we target very long periods over half a year. Therefore, we also change the price inputs to ticks from the best price.
	\item Deletion of time since previous order: On a tick-time scale, the order arrival interval should be modeled separately because some orders are published simultaneously, and their sequence and interval may not be significant. Thus, to simplify the problem we address in this study, we ignore this interval prediction.
\end{itemize}
The other architectures remain the same.

% たたかれる
% However, unlike other models, we set the total learning epochs of S-GAN to half of that for others (About the learning epoch, we have explained the learning epoch in the experiments section).
% This is because S-GAN employed gradient penalty \cite{Gulrajani2017} for 1-Lipschitz constraint.
% Gradient penalty requires an additional backpropagation for calculating the penalty in the loss.
% Thus, it requires more computational resources than others.

Moreover, we converted generated output to discrete values for fair evaluation by just rounding in the evaluation phase because the generated output is continuous numbers.

\subsubsection{\bf DCGAN}
As another comparative model, we employ a well-known DCGAN.
This model generates the order with continuous values similar to S-GAN.
The architecture is based on CNN.
Moreover, we also converted generated output to discrete values as the same as S-GAN.

\section{Experiments}
In our experiments,  we randomly selected 10 stocks under the criteria.
The basic premise is that we target only Tokyo Stock Exchange (TSE) in this study.
Thus, all stock candidates are listed on the TSE.

The first criterion is that the stocks must be included in Nikkei 225.
Nikkei 225 \footnote{\url{https://indexes.nikkei.co.jp/en/nkave/index/profile?idx=nk225}} is one of the major indices in TSE.
The stocks included in Nikkei 225 are selected in terms of their liquidity and sector balance, indicating that they have enough liquidity and are traded frequently.
In this study, we aim to generate realistic tick-scale orders.
Thus, liquidity is required.
The stocks included in this index are renewed when a stock is unlisted on TSE, and also periodically renewed every October.

The second criterion is that the stocks must not be included in TOPIX 100.
TOPIX 100 is also one of the major indices in TSE and selected by Japan Exchange Group \footnote{\url{https://www.jpx.co.jp/english/markets/indices/topix/}}.
TOPIX index series have some categories and some components.
TOPIX 100 includes the TOP-100 stocks whose total market value and liquidity are very high.
It is very special for our study because the stocks included in TOPIX 100 are treated as special stocks in terms of the trading rule.
These stocks have a smaller minimum order price unit (price tick size).
Combined with the fact that the stocks are chosen for their high liquidity, these stocks result in a very high trading volume.
It is very challenging for us to choose these stocks in terms of computational resources.
Thus, we decide to ignore those included in TOPIX 100.

The third criterion is that the stocks are included in TOPIX 225, but not included TOPIX 100, stably in 2018 -- 2020 period.
It is because acceptance or deletion by indices have a significant impact on trading volume.

The last criterion is that the stocks have the same price tick size through the data periods.
In TSE, the price tick size changes according to the price range.
It changes at the price of $N \times 10^M (N = 1, 3, 5, M = 3, 4, 5, 6, 7)$.
Due to the current technological problem, our model cannot accept the change in price tick size.
Thus, we decided to employ this criterion.

\begin{table*}[htb]
	\centering%\small
	%\resizebox{.95\columnwidth}{!}{
	\caption{Selected Stocks}
	\begin{tabular}{ccccc}
		\hline
		Ticker  & Name                             & Classification (by Bloomberg)     & \# of orders in data \\\hline
		5901 JP & Toyo Seikan Group Holdings, Ltd. & Containers \& Packaging        & 6,569,563            \\
		5333 JP & NGK Insulators, Ltd.             & Auto Parts                     & 6,077,554            \\
		8355 JP & Shizuoka Bank, Ltd.              & Banks                          & 5,307,488            \\
		5631 JP & Japan Steel Works, Ltd.          & Other Machinery \& Equipment   & 6,787,814            \\
		9532 JP & Osaka Gas Co., Ltd.              & Gas Utilities                  & 7,914,464            \\
		7012 JP & Kawasaki Heavy Industries        & Diversified Industrials        & 9,122,778            \\
		2501 JP & Sapporo Holdings, Ltd.           & Alcoholic Beverages            & 4,852,475            \\
		4005 JP & Sumitomo Chemical Co., Ltd.      & Basic \& Diversified Chemicals & 6,319,126            \\
		7752 JP & Richo Co. Ltd.                   & Consumer Electronics           & 6,942,513            \\
		7911 JP & Toppan Inc.                      & Printing Services              & 6,057,922            \\
		\hline
	\end{tabular}
	\label{table:stocks}
\end{table*}

As a data period, we employed January -- September in 2019, because we avoid the periodical updates of indices.
Nikkei 225 is periodically renewed on the every first business day of August, and TOPIX 100 on every last business day of August.

According to these criteria, we obtain 81 stocks.
Only 125 stocks are included in Nikkei 225 and not in TOPIX 100.
Thus, candidates for random selection are more than half of the stock candidates.

From these 81 stocks, we choose 10 at random, which is shown in table \ref{table:stocks}.

The data are split as train:valid:test = 8:1:1 in temporal sequence.

The test task is the next order generation.
The generation of a long time series is also a repetition of the prediction of the next order.
For simplification, we narrow it down to the generation of the next order.
% This is partially reasonable because there has been no research on generating only long order time series, but it should be addressed in future work.
For PGSGAN, we calculate and inspect the negative log-likelihood (NLL) for real order and the entropy of the generated policy.
The NLL is $\mathrm{NLL}_{G(z)} (x)$ where $x$ is the real order.
This shows how the generator policy successfully fits the real order.
Although a low NLL indicates a better fit with the real data, a complete fit is not beneficial as a generator.
Thus, our experiments use the log-likelihood as one index for check learning status, but do not pursue the lowest NLL.
Theoretically, the by-chance NLL is $- \ln{\{1 / (2 \times 2 \times 2 \times 40 \times 40)\}} \approx 9.45$.
On the contrast, the entropy is defined as
\begin{eqnarray}
	H(G(z)) := \sum\nolimits_{x\in \mathbb{X}} - p_{G(z)}(x)\log_2{p_{G(z)}(x)},
\end{eqnarray}
where $\mathbb{X}$ is all order classes.
This entropy indicates how well the generator policy is learned.
If the policy is learned well, the probability of each class of the generated policy will be well skewed.
Therefore, this index is useful for checking the convergence of PGSGAN.
Theoretically, the by-chance entropy is
\begin{eqnarray}
\sum_{x\in \mathbb{X}} - \frac{1}{2 \times 2 \times 2 \times 40 \times 40}\log_2{\frac{1}{2 \times 2 \times 2 \times 40 \times 40}}\\
= \log_2{(2 \times 2 \times 2 \times 40 \times 40)}
\approx 13.64.
\end{eqnarray}
where $\mathbb{X}$ is all order class.

Moreover, we also compare the distribution of generated and real orders, for all the orders.
Here, we employ Kullback--Leibler divergence (KLD) \cite{kullback1951} and Mean Square Error (MSE) for all classes ($2 \times 2 \times 2 \times 40 \times 40$).
Kullback--Leibler divergence is defined as:
\begin{eqnarray}
	D_{KL}(P,Q) = \sum\nolimits_{x\in \mathbb{X}} P(x)\log_2{\left({P(x)}/{Q(x)}\right)},
\end{eqnarray}
where $\mathbb{X}$ is all order class, and $P(x)$ and $Q(x)$ indicate the probability of the real orders and the generated orders for class $x$, respectively.
Even though $P(x)\log_2{\left({P(x)}/{Q(x)}\right)}$ is calculated as 0 when $P(x) = (Q(x) =)~ 0$, KLD would be infinity if $\exists x, P(x) \neq 0, Q(x) = 0$ due to some reasons, such as mode collapse of generator.
Moreover, for DCGAN and S-GAN, because the generated output is continuous numbers, we round the output to translate into the discreate values.
Because the generated output should be different based on random seeds, we evaluate the generator 100 times with different seeds in each situation in test data.

As experiments settings, we employ a batch size of 2048, 5000 epochs maximum, the learning rate (both the generator and critic) of $10^{-5}$, and the Adam optimizer.
Moreover, the balance of learning chance of the generator and critic (two time-scale update rule \cite{heusel2017gans}) is set to $1:5$.
% In addition, to improve computational efficiency, models are saved for every 10 epochs, and only those models could be used for tests.

\section{Results}
\begin{table*}[tb]
	\centering%\small
	%\resizebox{.95\columnwidth}{!}{
	\caption{All the results of KLD}
	% \vspace{-2mm}
	\begin{tabular}{c|cccc}
		\hline
		Ticker  & PGSGAN (KLD)                   & PGSGAN-HL (KLD)                & S-GAN (KLD)             & DCGAN (KLD)             \\\hline\hline
		5901 JP & $0.208467\pm0.000107$          & $\mathbf{0.168823\pm0.000128}$ & $\infty\pm\mathrm{NaN}$ & $\infty\pm\mathrm{NaN}$ \\\hline
		5333 JP & $0.257479\pm0.000121$          & $\mathbf{0.203760\pm0.000121}$ & $\infty\pm\mathrm{NaN}$ & $\infty\pm\mathrm{NaN}$ \\\hline
		8355 JP & $\mathbf{0.136612\pm0.000071}$ & $0.139955\pm0.000061$          & $\infty\pm\mathrm{NaN}$ & $\infty\pm\mathrm{NaN}$ \\\hline
		5631 JP & $0.215376\pm0.000064$          & $\mathbf{0.209126\pm0.000077}$ & $\infty\pm\mathrm{NaN}$ & $\infty\pm\mathrm{NaN}$ \\\hline
		9532 JP & $0.198947\pm0.000053$          & $\mathbf{0.190806\pm0.000078}$ & $\infty\pm\mathrm{NaN}$ & $\infty\pm\mathrm{NaN}$ \\\hline
		7012 JP & $0.150013\pm0.000067$          & $\mathbf{0.138801\pm0.000096}$ & $\infty\pm\mathrm{NaN}$ & $\infty\pm\mathrm{NaN}$ \\\hline
		2501 JP & $0.243355\pm0.000108$          & $\mathbf{0.216768\pm0.000173}$ & $\infty\pm\mathrm{NaN}$ & $\infty\pm\mathrm{NaN}$ \\\hline
		4005 JP & $\mathbf{0.136136\pm0.000053}$ & $0.144642\pm0.000071$          & $\infty\pm\mathrm{NaN}$ & $\infty\pm\mathrm{NaN}$ \\\hline
		7752 JP & $0.164498\pm0.000036$          & $\mathbf{0.123295\pm0.000046}$ & $\infty\pm\mathrm{NaN}$ & $\infty\pm\mathrm{NaN}$ \\\hline
		7911 JP & $0.228341\pm0.000083$          & $\mathbf{0.203080\pm0.000107}$ & $\infty\pm\mathrm{NaN}$ & $\infty\pm\mathrm{NaN}$ \\\hline
	\end{tabular}
	\label{table:KLD}
\end{table*}
\begin{table*}
	% \vspace{-1mm}
	\centering%\small
	%\resizebox{.95\columnwidth}{!}{
	\caption{All the results of MSE}
	% \vspace{-2mm}
	\begin{tabular}{c|cccc}
		\hline
		Ticker  & PGSGAN (MSE)          & PGSGAN-HL (MSE)                & S-GAN (MSE)           & DCGAN (MSE)           \\\hline
		5901 JP & $0.000479\pm0.000003$ & $\mathbf{0.000386\pm0.000003}$ & $0.019759\pm0.000053$ & $0.148968\pm0.000005$ \\\hline
		5333 JP & $0.002164\pm0.000002$ & $\mathbf{0.000945\pm0.000005}$ & $0.004611\pm0.000011$ & $0.143241\pm0.000007$ \\\hline
		8355 JP & $0.000505\pm0.000002$ & $\mathbf{0.000392\pm0.000001}$ & $0.012347\pm0.000018$ & $0.093950\pm0.000016$ \\\hline
		5631 JP & $0.000499\pm0.000002$ & $\mathbf{0.000432\pm0.000003}$ & $0.027200\pm0.000035$ & $0.107452\pm0.000026$ \\\hline
		9532 JP & $0.000300\pm0.000000$ & $\mathbf{0.000278\pm0.000003}$ & $0.030383\pm0.000029$ & $0.145481\pm0.000011$ \\\hline
		7012 JP & $0.000332\pm0.000003$ & $\mathbf{0.000257\pm0.000003}$ & $0.020080\pm0.000042$ & $0.145100\pm0.000003$ \\\hline
		2501 JP & $0.000428\pm0.000002$ & $\mathbf{0.000391\pm0.000003}$ & $0.014947\pm0.000039$ & $0.146345\pm0.000012$ \\\hline
		4005 JP & $0.000512\pm0.000004$ & $\mathbf{0.000393\pm0.000003}$ & $0.014715\pm0.000017$ & $0.109629\pm0.000009$ \\\hline
		7752 JP & $0.000866\pm0.000003$ & $\mathbf{0.000691\pm0.000004}$ & $0.009458\pm0.000015$ & $0.141400\pm0.000005$ \\\hline
		7911 JP & $0.000580\pm0.000003$ & $\mathbf{0.000473\pm0.000004}$ & $0.019275\pm0.000029$ & $0.107608\pm0.000012$ \\\hline
	\end{tabular}
	\label{table:MSE}
	% \vspace{-2mm}
\end{table*}

Tables \ref{table:KLD} and \ref{table:MSE} show all the results of KLD and MSE, respectively, between fake and real distributions.
Each row shows a randomly selected ticker (listed company).
The best performances for each ticker are written in bold.
S-GAN and DCGAN have no finite KLD by the reason mentioned above.

In terms of KLD, the performances of PGSGAN and PGSGAN-HL depend on tickers.
However, our proposed models outperform others.
All models successfully have MSE measures.
According to the results of MSE, our PGSGAN-HL shows the best performances in all the selected tickers.
Moreover, PGSGAN also outperforms S-GAN and DCGAN in all tickers.

As shown in previous studies, the S-GAN outperformed the DCGAN.
However, compared with our proposed model, the performances of S-GAN are very limited.

\begin{figure*}[htb]
	\centering\footnotesize
	PGGAN 5901 JP\\
	\includegraphics[width=\linewidth]{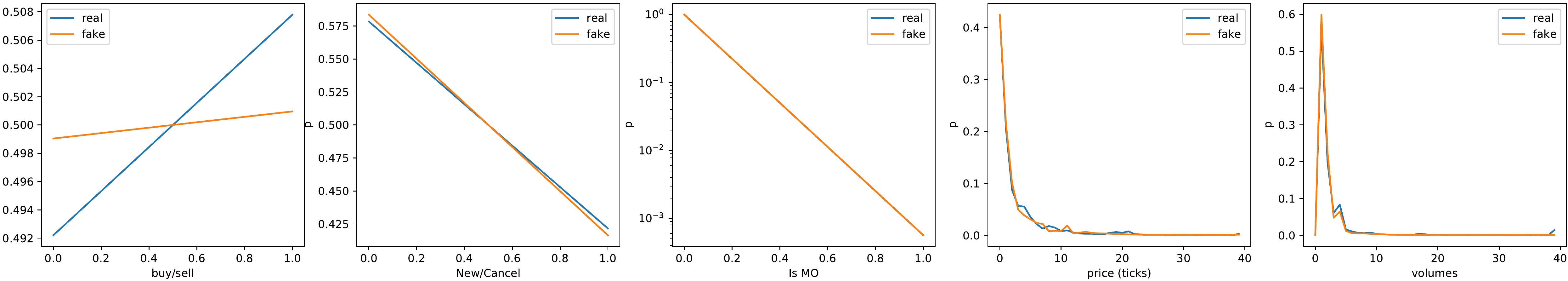}
	PGGAN-HL 5901 JP\\
	\includegraphics[width=\linewidth]{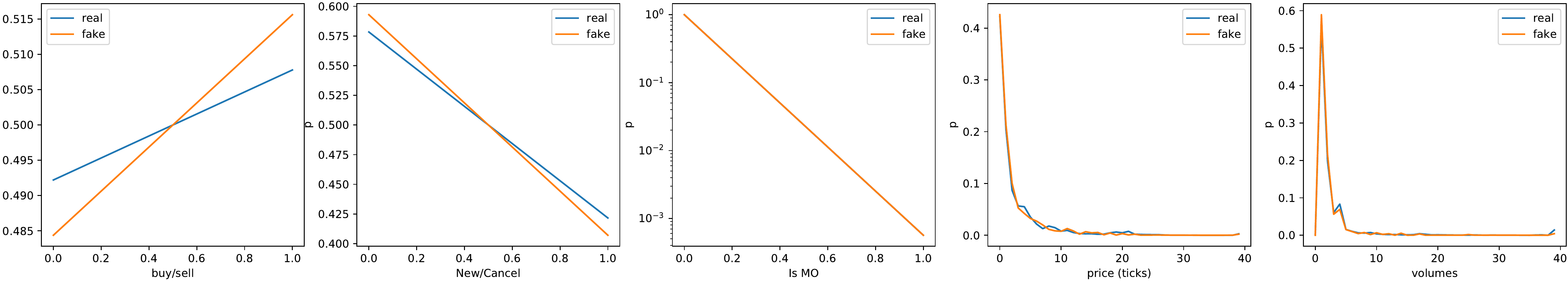}
	S-GAN 5901 JP\\
	\includegraphics[width=\linewidth]{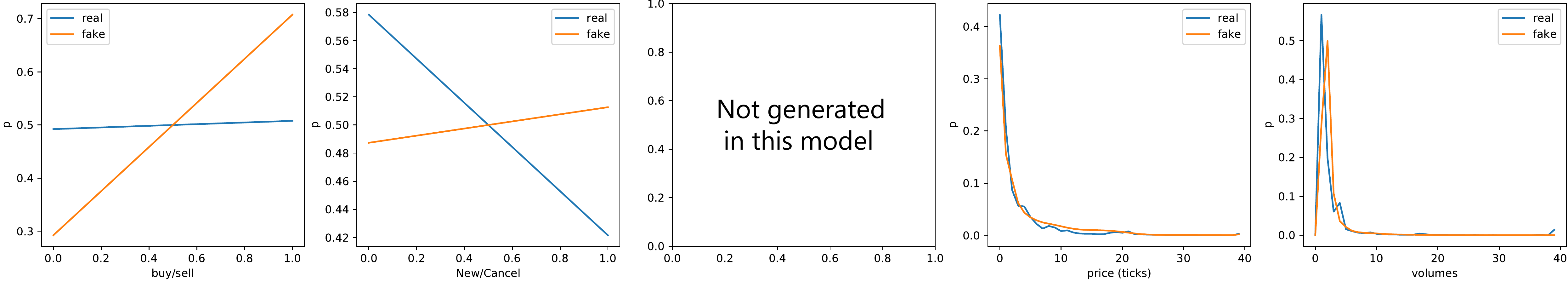}
	DCGAN 5901 JP
	\includegraphics[width=\linewidth]{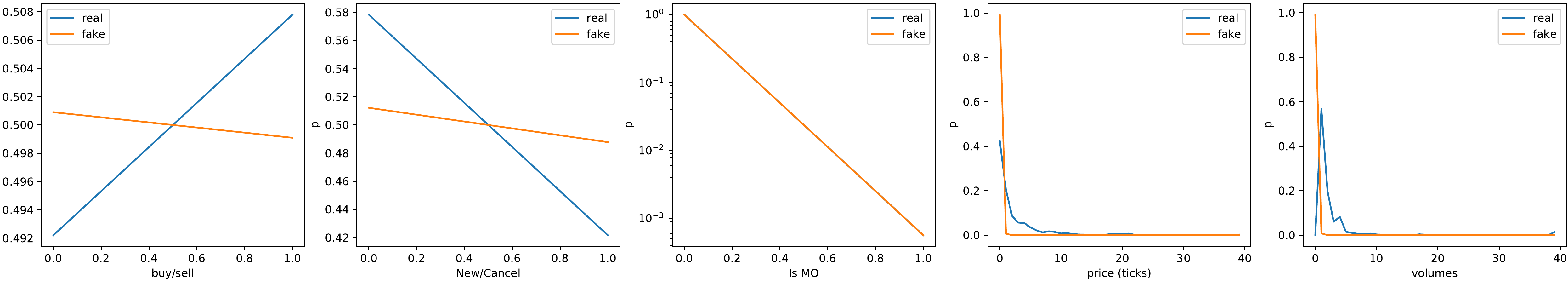}
	\vspace{-6mm}
	\caption{An example of the distribution of real (blue) and generated fake (orange) data.
		% The bigger version is shown in the supplementary material.
		From left to right, each box shows the distribution of sell/buy, new/cancel, MO (or not), price (ticks from the opposite best price), and volume (divided by the minimum volume).
		In the leftmost box, 0.0 and 1.0 in the horizontal axis correspond to sell and buy, respectively.
		In the second left box, 0.0 and 1.0 correspond to new and cancel, respectively.
		In the middle box, 0.0 and 1.0 correspond to non-MO and MO, respectively.
		In the vertical direction, only the model is different; therefore, the real data (blue) is the same; only the vertical axis scale is different.}
	\vspace{-3mm}
	\label{fig:fake-ex}
\end{figure*}

To further inspect details of the distribution comparison, as shown in figure \ref{fig:fake-ex}, we also make detailed figures of the generated orders' distributions.
Here, we only show one example from 5901 JP. %, and the figure is the small version.
Those for other tickers, are presented in appendix.

According to the figure, the failure in reproducing the volume and price distributions in DCGAN is notable.

Further, the failure in reproducing the tiny bumps in the price and volume distributions of S-GAN is interesting.
Unlike PGSGAN/PGSGAN-HL, S-GAN has smoother distributions of price and volumes, which is the bigger difference from the real distribution than PGSGAN/PGSGAN-HL
% (This detail can be better observed in the bigger version of this figure in the supplementary material).

\section{Discussion}
Our models surpass that of previous studies.
PGSGAN-HL and PGSGAN outperforming the previous works implies that our implementation of policy gradient is beneficial for GANs for stock markets.

Under the given rules of the financial markets, mapping to a discrete space is more reasonable than mapping to a continuous space.
Of course, the price or volumes can be placed on the continuous space.
However, possible order space is completely discrete, even in price and volume.

For that implementation of discrete space, we employ the policy gradient, to fill in the disconnection of the gradient between the generator and the critic.
As shown in the theoretical discussion, we have also succeeded in incorporating the policy gradient into the GAN in experiments.

\begin{figure*}[htbp]
	\centering
	\includegraphics[width=0.7\linewidth]{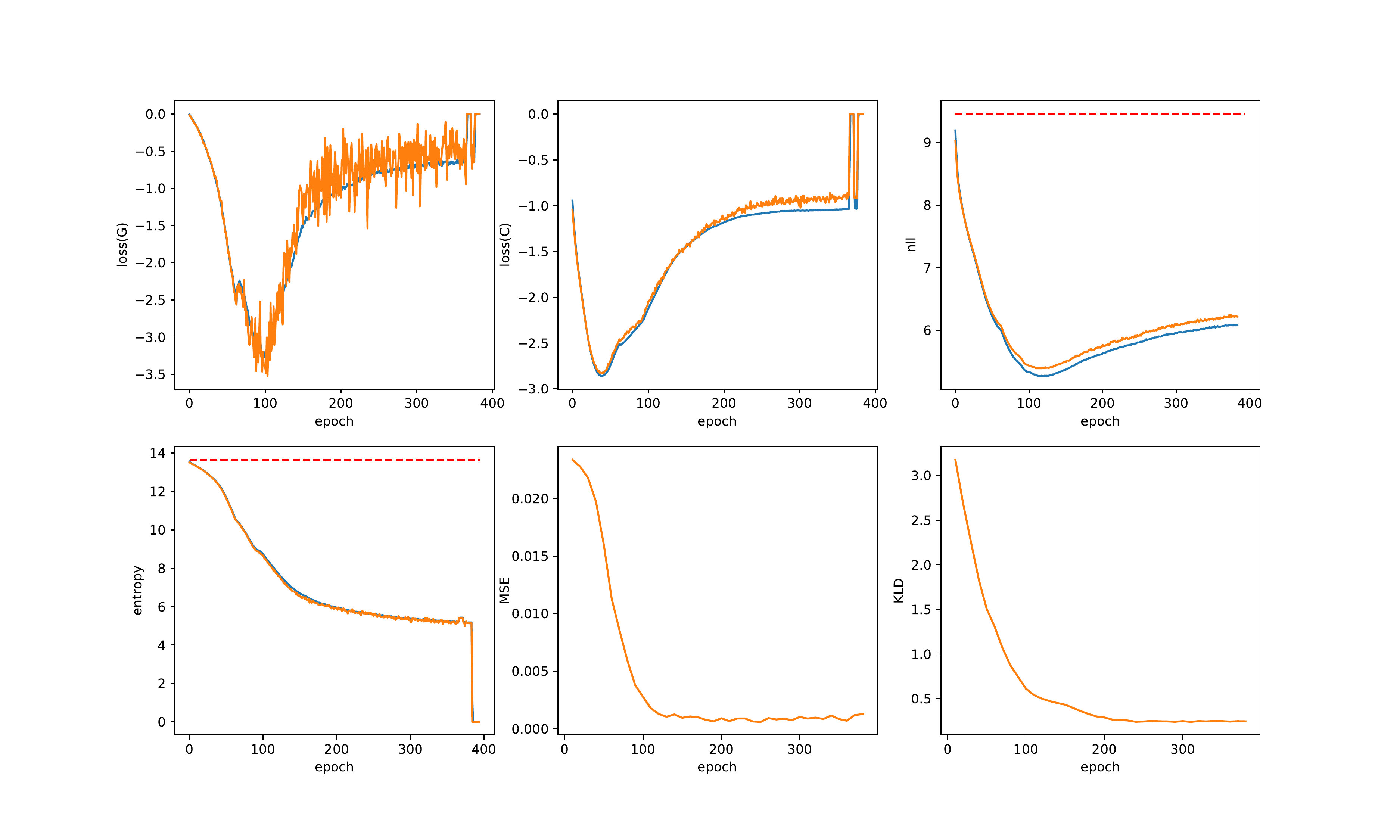}
	\vspace{-3mm}
	\caption{An example of the metrics in the learning process. (PGSGAN 5901 JP)
		From upper left to upper right, each box shows the loss of generator, loss of critic, and NLL ($\mathrm{NLL}_{G(z)} (x)$ where $x$ is the real order), respectively.
		From lower left to lower right, each box shows the entropy of generated policy, the MSE between real and fake distributions, and the KLD between them, respectively.
		Horizontal axis represents epochs.
		Blue and orange lines correspond to train and valid results, respectively.
		The MSE and KLD are calculated only in the valid data.
		Moreover, a broken red line means the by-chance level explained previously.}
	\label{fig:learning}
\end{figure*}

As explained, generated policy entropy can be used to monitor the current learning status.
To validate this by experiments, we plot each metric during the learning process of PGSGAN in figure \ref{fig:learning}.
Obviously, the losses are not beneficial for monitoring the learning status because the generator and the critic are adversarial, and their losses are merely relative.
This resembles other usual GANs, which usually employ the outer task to monitor the status.
For an example of image generations, the inception score \cite{Salimans2016} and Fréchet Inception Distance \cite{heusel2017gans} are introduced to evaluate the current learning level.
However, setting these outer tasks for evaluating the learning status is complicated, especially in financial markets.
According to the result in figure \ref{fig:learning}, the plot of the entropy is very similar to the MSE and KLD results.
Thus, our experiments show that the entropy of generated policy is also beneficial.

Whereas the generated distribution is similar to the real distribution, this is a necessary but not a sufficient condition for a good generator.
As we showed and explained in the results, our PGSGAN has a better fake distribution similar to the real distribution than other models.
However, there is a good possibility of mode collapse occurring, deprecating only the real orders, or generating fake orders according to the real distribution, while ignoring the situation.

We also tested this by changing fake seeds 100 times in the test.
If the generator has enough capability to make various fake orders, changing the seeds results in the different fake orders.

To check if the generator can make various appropriate fake orders, we employ NLL ($\mathrm{NLL}_{G(z)} (x)$ where $x$ is the real order) as the metric.
If the NLL is almost the same as the by-chance level, the generator would make only meaningless fake orders.
By contrast, if the NLL is 0, the generated would make only the real orders and fail to make various likely fake orders.
Moreover, if the NLL for one situation is always the same, not depending on the seeds, it would have a high possibility of mode collapse of generated policy.
Thus, we calculate the mean of the standard deviation of NLL on each generating situation by changing seeds 100 times.
This evaluation is only possible for PGSGAN/PGSGAN-HL.
Thus, we find that PGSGAN/PGSGAN-HL shows the appropriate level of a variety of generations.
The standard deviation of NLL for each generation by changing seeds is roughly 2.5--3.6 on average.
Because the mean of NLL is roughly 5--6.5, the deviation of NLL is moderate enough.
Moreover, the NLL is also moderate level.
This result indicates that our proposed model has a sufficient variety of generations and mode collapses do not occur.

In summary, our model fulfills the necessary and sufficient conditions for good generation at the least level.
Although there is a possibility of making a better model, we can say that our model fulfills the requirements of a good generator and performs better than the previous ones.

Lastly, the reason PGSGAN-HL showed better performances in many experiments than PGSGAN is the existence of Hinge loss.
Different from PGSGAN-HL, the gradient of the generator of PGSGAN will be almost 0 in the middle of learning.
As described earlier in equation \ref{eq:loss_g}, the gradient of the generator depends on and is proportional to the output of the critic.
If the generated fake data is more perfect than the critic's discrimination, the critic's output will be 0 (See figure \ref{fig:learning}).
This causes the vanishing of the gradient in the generator and the explicit end of the learning of PGSGAN.
By introducing Hinge loss, this can be avoided.
Moreover, the learning can be continued for a longer time than PGSGAN.

As future work, higher performance GANs, evaluation methods better than MSE/KLD, or application of our GANs, can be pursued.
We have only focused on the next orders' generation to simplify the problem.
However, the challenge to make longer time series needs to be addressed.

\section{Conclusion}
We proposed a new GAN for generating realistic orders in financial markets.
GANs in some previous works generated fake orders in continuous spaces because of GAN architectures' learning limitations.
However, the real orders are discrete.
For example, the price and volumes have minimum units.
Moreover, order types, such as sell/buy, are also not continuous; it is inappropriate to join their state spaces continuously.
Thus, in this study, we change the generated fake orders to discrete orders.
Because this change disabled the ordinary GAN learning algorithm, this study newly employed policy gradient for the learning algorithm.
Policy gradient is frequently used in reinforcement learning.
In this study, we made it possible to use policy gradients by incorporating the relationship between the generator and the critic into the reinforcement learning framework.
In our model, the generator makes a policy; then, according to the policy, randomly sampled fake orders are processed by the critic.
Our experiments tested our models, policy gradient stock GAN (PGSGAN) and policy gradient stock GAN with Hinge loss (PGSGAN-HL), in terms of next order generations.
The data we used in this study were the order data from TSE.
Then, we compared the generated fake orders' distribution and the real order distribution in terms of their MSE and KLD.
As a result, we demonstrated that our proposed model outperforms the previous ones.
In addition, as a side benefit of introducing the policy gradient, we found that the entropy of the generated policy can be used to check the learning status of the GAN.
Moreover, the combination of our model and Hinge loss (PGSGAN-HL) seems to be beneficial for better learning by avoiding the gradient vanishing.
As future work, higher performance GANs, evaluation methods better than MSE/KLD, or application of our GANs, should be addressed.

\section*{Acknowledgment}
We thank the Japan Exchange Group, Inc. for providing the data.
This work was supported by JSPS KAKENHI Grant Number JP 21J20074 (Grant-in-Aid for JSPS Fellows).
% \footnotesize
\bibliographystyle{IEEEtrans}
\bibliography{cite}

\appendix

\subsection{Additional Note for Data}
% Table \ref{table:stocks} shows all the 10 selected stocks.
% It also contains the number of orders in the data.

Compared with Stock-GAN (S-GAN) \cite{Li2020}, the data size is huge.
There are two selected stocks in the experiment of S-GAN -- Alphabet Inc. (GOOG) and Patriot National Bancorp Inc. (PNBK).
In the study, GOOG had 230,000 orders, whereas PNBK had only 20,000 orders.
S-GAN was tested for only almost one day.
However, in this study, longer periods (9 months) and bigger data are tested.

\subsection{Calculation Time for Learning}
Here, we discuss the learning time for each model.
In the settings of our experiments, each model took the following learning time:
\begin{itemize}
	\item PGSGAN: 3--5 days.
	\item PGSGAN-HL: About 20 days.
	\item S-GAN: About 6 months% (Estimation. About 3 months for half of the total learning of the same epochs of other models.)
	\item DCGAN: About 10 days.
\end{itemize}
These times are roughly calculated because they depend on the size of the data (stocks) and computational resources.
Thus, whereas this is not accurate, it is beneficial for understanding the learning difficulty of each model.
Additionally, these values are measured with high-end GPUs of NVIDIA Geforce RTX 20 series, such as 2080, 2070super, and 2080Ti.

Certainly, S-GAN has the greatest learning difficulty.
The biggest problem of S-GAN is the gradient penalty \cite{Gulrajani2017} for 1-Lipschitz constraint.
% As mentioned in the main paper, this requires one additional backpropagation to calculate the loss's penalty.
Roughly, the gradient penalty makes the calculation cost twice.
However, even if the learning time is halved, it is still long.
Another problem of S-GAN is the insufficient setting of convolutional layers.
The kernel size of the convolutional layers is too big, and it does not take the full advantage of CNN.

The reason that PGSGAN's learning time is short is the explicit end of the learning.
Different from PGSGAN-HL, the gradient of the generator of PGSGAN will be almost 0 in the middle of learning.
As described in main contents, the gradient of the generator depends on and remains proportional to the output of the critic.
If the generated fake data is more perfect than the critic's discrimination, the critic's output will be 0.
This causes the gradient in the generator to vanish.

\subsection{All Result Graphs}
\begin{figure*}[htbp]
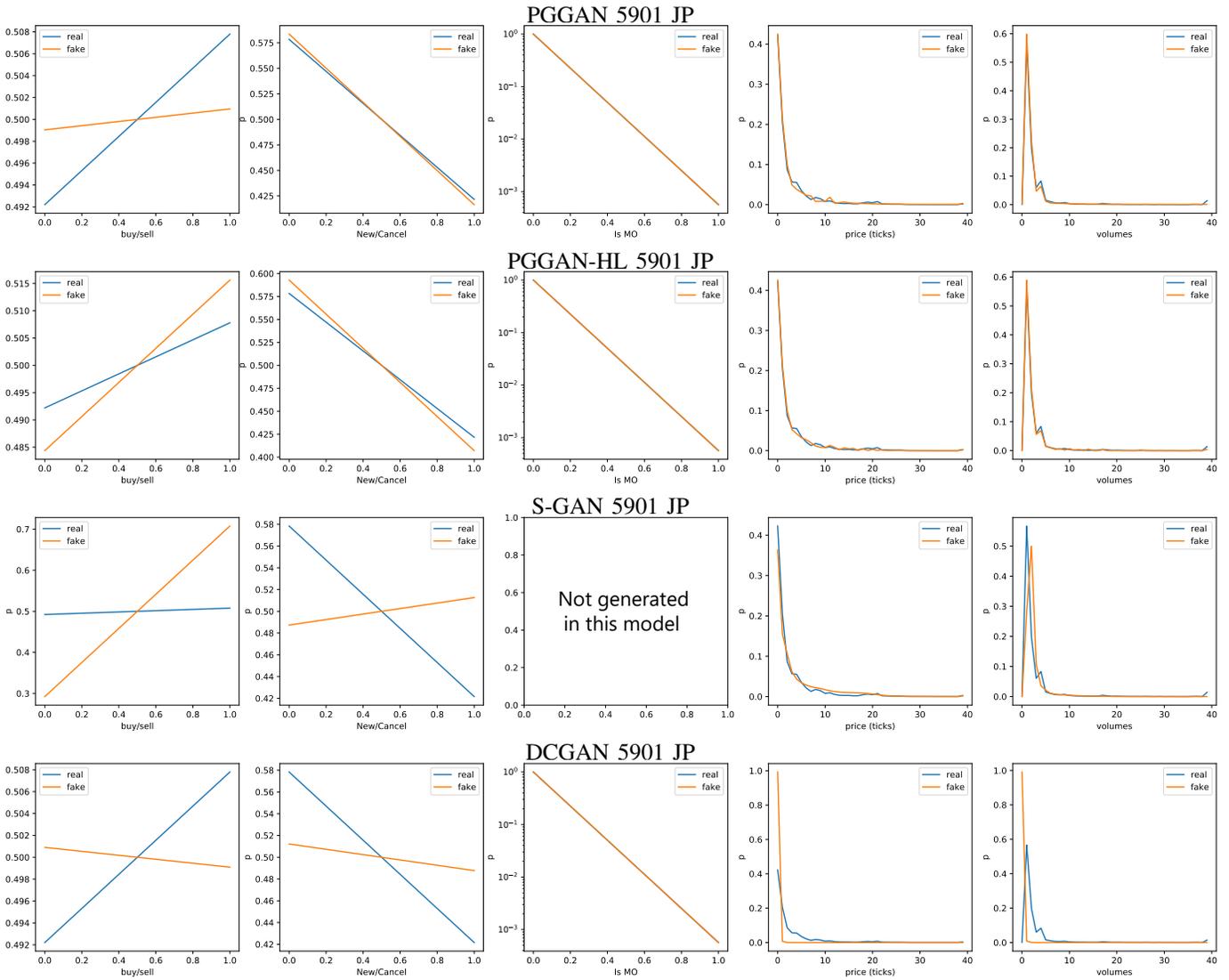

	\centering
	PGGAN 5901 JP\\
	\includegraphics[width=\linewidth]{figs/PGSGAN-5901-0000300-test2.pdf}
	PGGAN-HL 5901 JP\\
	\includegraphics[width=\linewidth]{figs/PGSGAN-HL-5901-0004200-test2.pdf}
	S-GAN 5901 JP\\
	\includegraphics[width=\linewidth]{figs/S-GAN-5901-0000380-test.pdf}
	DCGAN 5901 JP
	\includegraphics[width=\linewidth]{figs/DCGAN-5901-0000010-test.pdf}
	\caption{The results of real and fake data distribution from each models in 5901 JP.}
	\label{fig:fake-5901}
\end{figure*}

\begin{figure*}[htbp]
	\centering
	PGGAN 5333 JP\\
	\includegraphics[width=\linewidth]{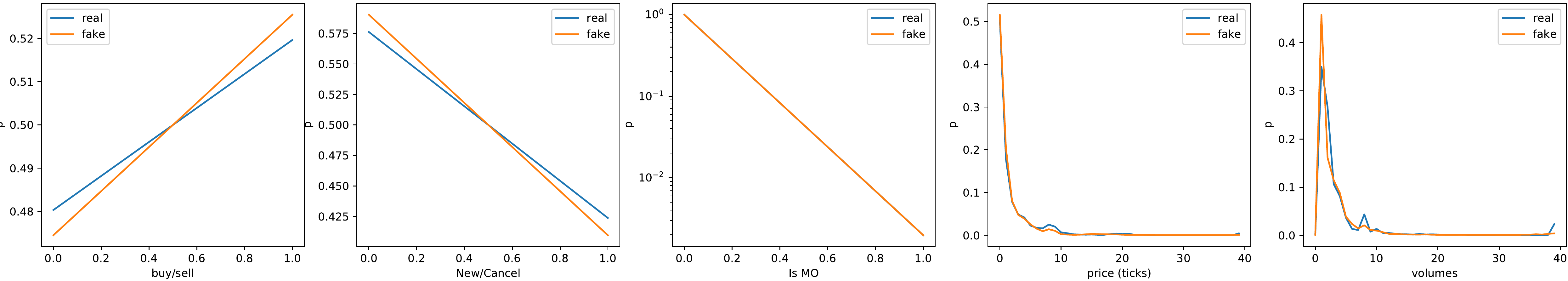}
	PGGAN-HL 5333 JP\\
	\includegraphics[width=\linewidth]{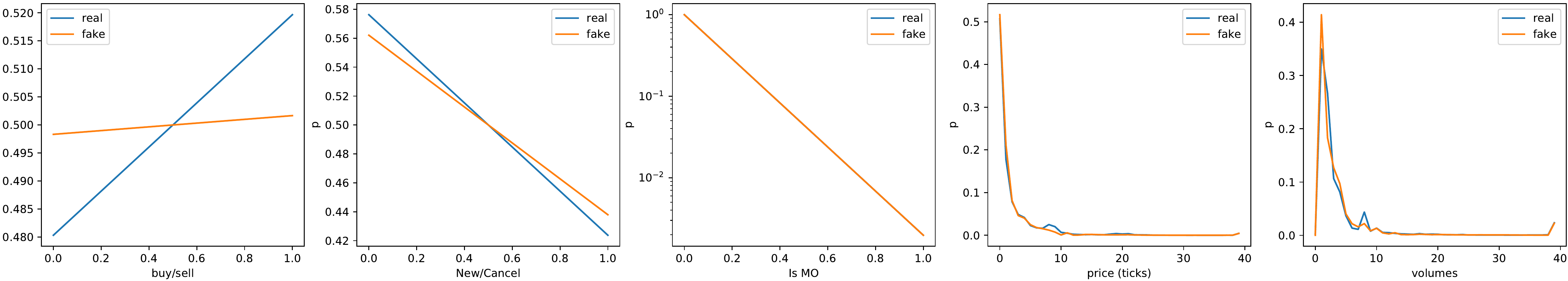}
	S-GAN 5333 JP\\
	\includegraphics[width=\linewidth]{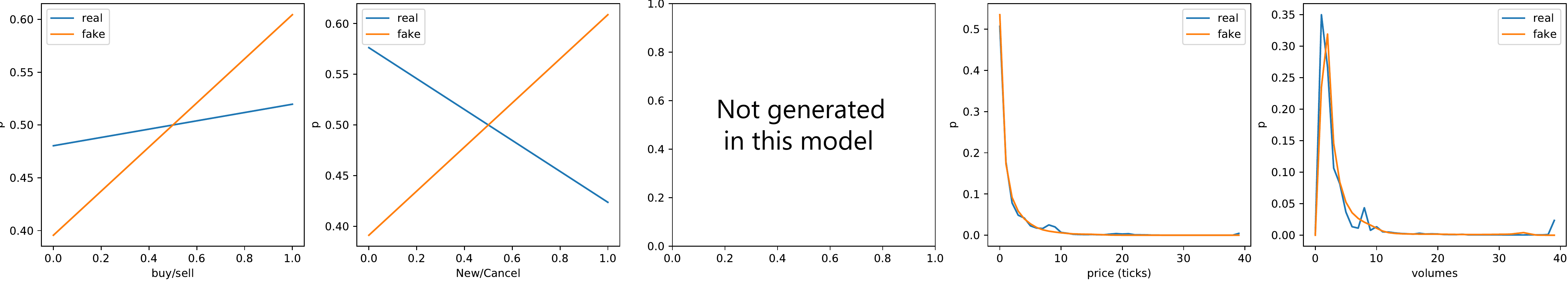}
	DCGAN 5333 JP
	\includegraphics[width=\linewidth]{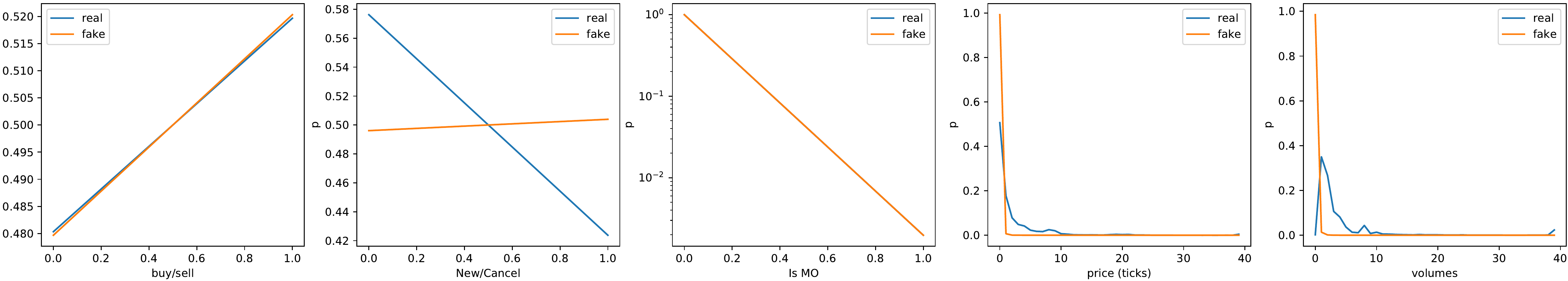}
	\caption{The results of real and fake data distribution from each models in 5333 JP.}
	\label{fig:fake-5333}
\end{figure*}

\begin{figure*}[htbp]
	\centering
	PGGAN 8355 JP\\
	\includegraphics[width=\linewidth]{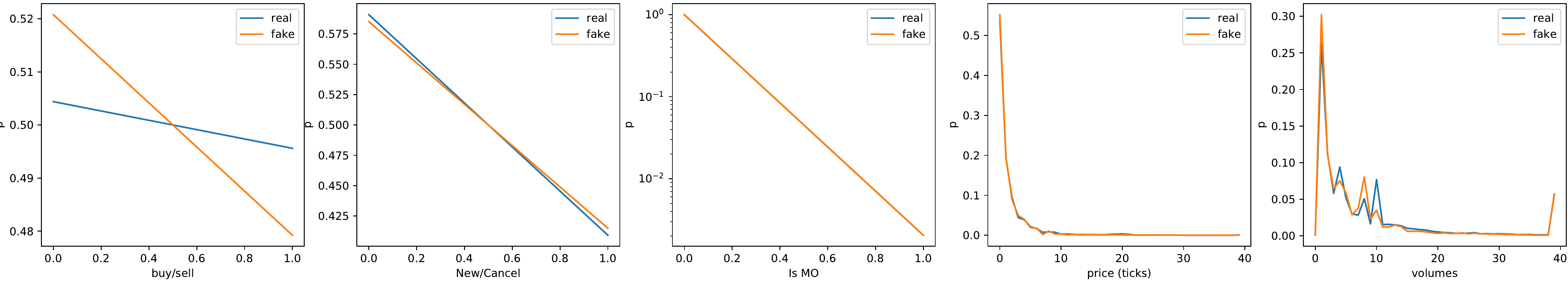}
	PGGAN-HL 8355 JP\\
	\includegraphics[width=\linewidth]{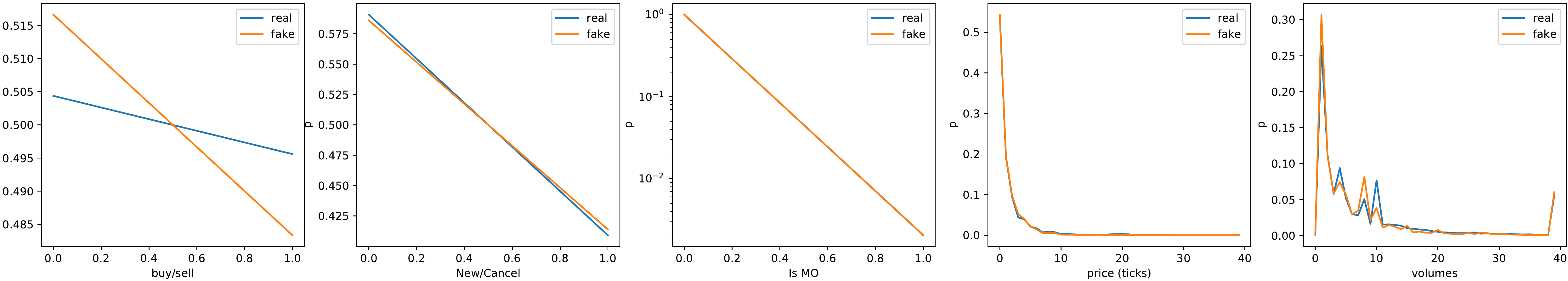}
	S-GAN 8355 JP\\
	\includegraphics[width=\linewidth]{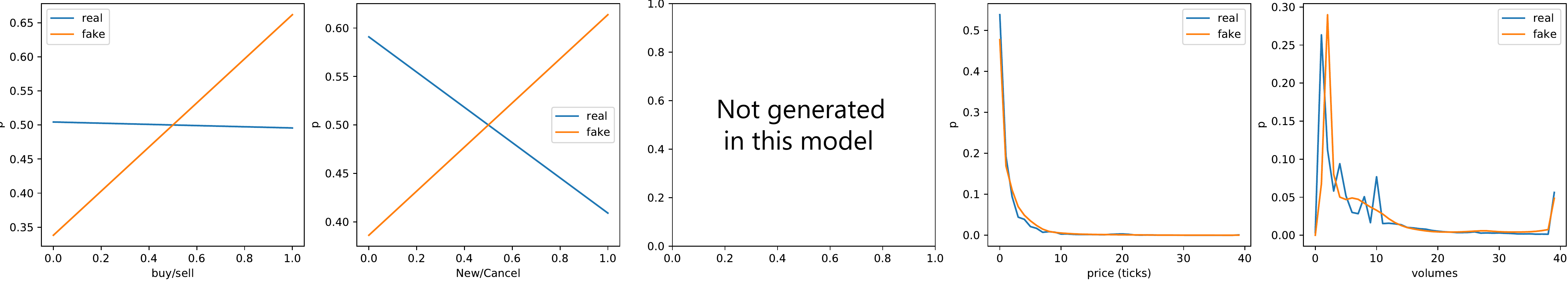}
	DCGAN 8355 JP
	\includegraphics[width=\linewidth]{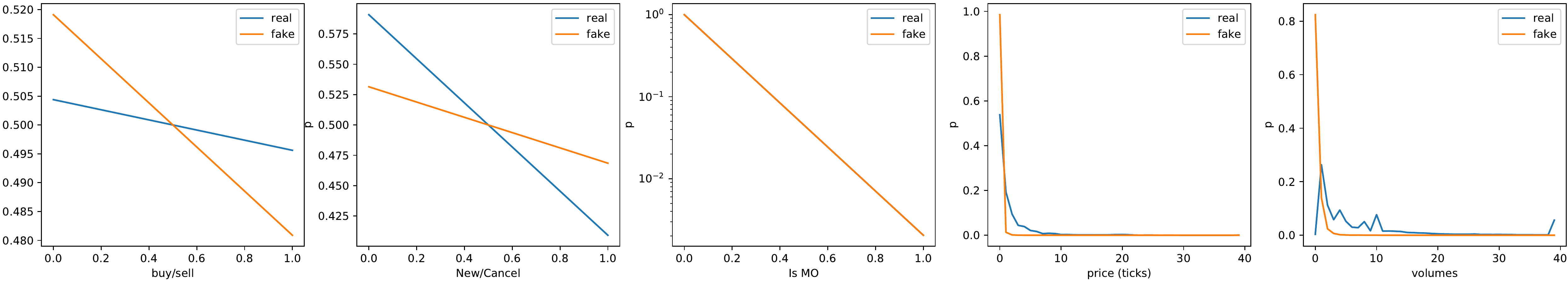}
	\caption{The results of real and fake data distribution from each models in 8355 JP.}
	\label{fig:fake-8355}
\end{figure*}

\begin{figure*}[htbp]
	\centering
	PGGAN 5631 JP\\
	\includegraphics[width=\linewidth]{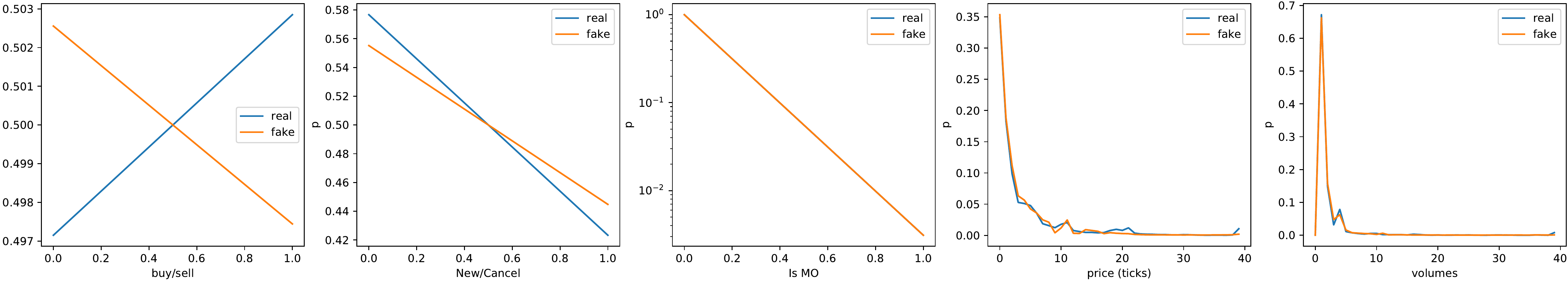}
	PGGAN-HL 5631 JP\\
	\includegraphics[width=\linewidth]{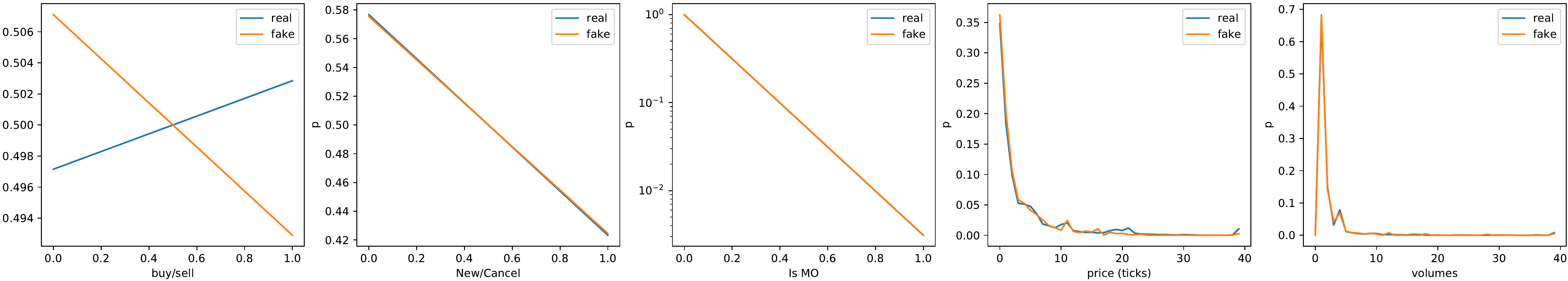}
	S-GAN 5631 JP\\
	\includegraphics[width=\linewidth]{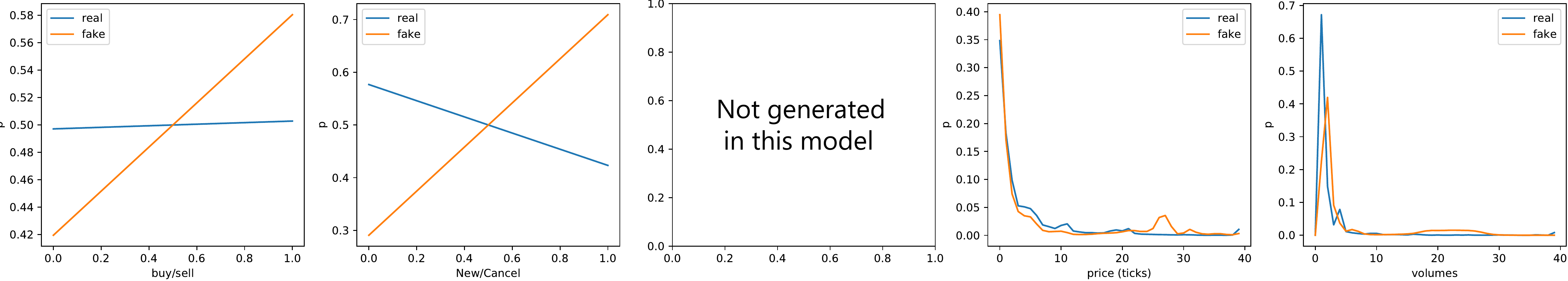}
	DCGAN 5631 JP
	\includegraphics[width=\linewidth]{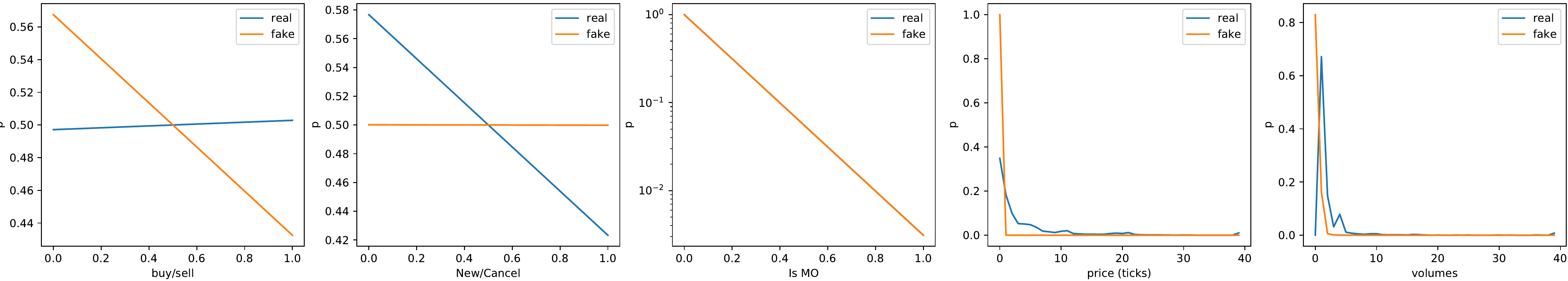}
	\caption{The results of real and fake data distribution from each models in 5631 JP.}
	\label{fig:fake-5631}
\end{figure*}

\begin{figure*}[htbp]
	\centering
	PGGAN 9532 JP\\
	\includegraphics[width=\linewidth]{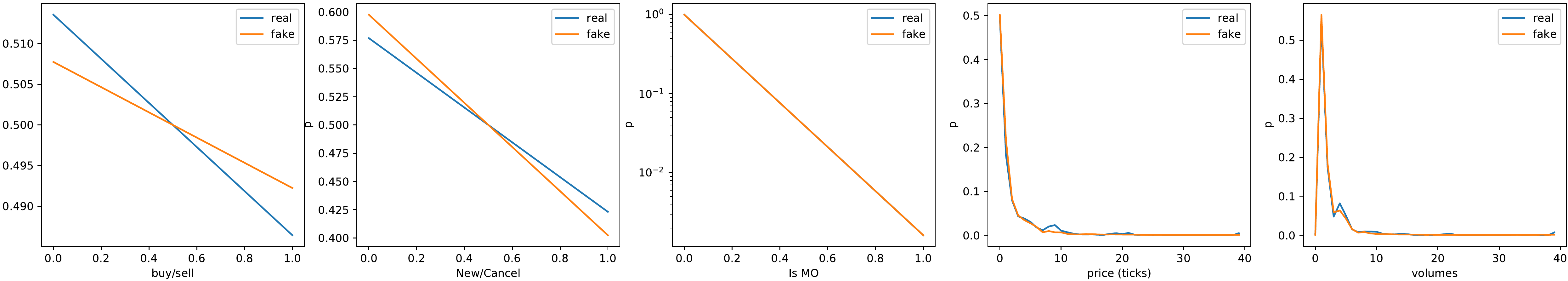}
	PGGAN-HL 9532 JP\\
	\includegraphics[width=\linewidth]{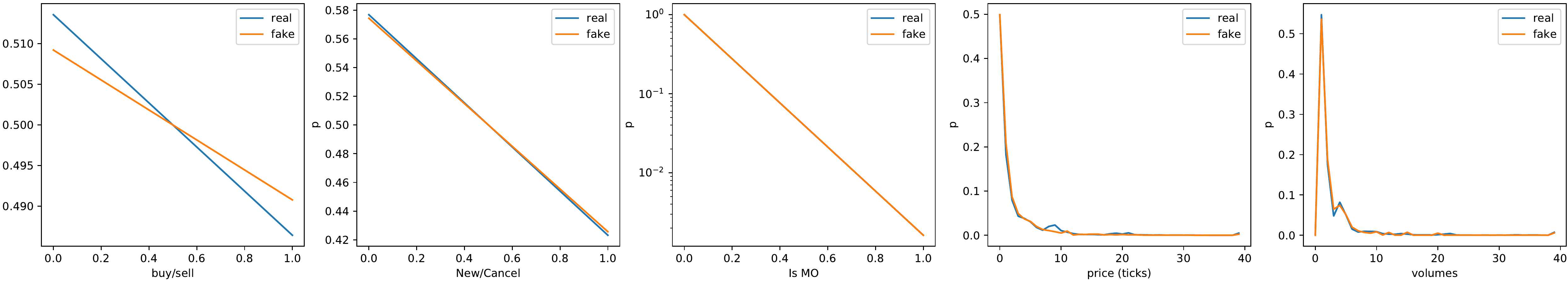}
	S-GAN 9532 JP\\
	\includegraphics[width=\linewidth]{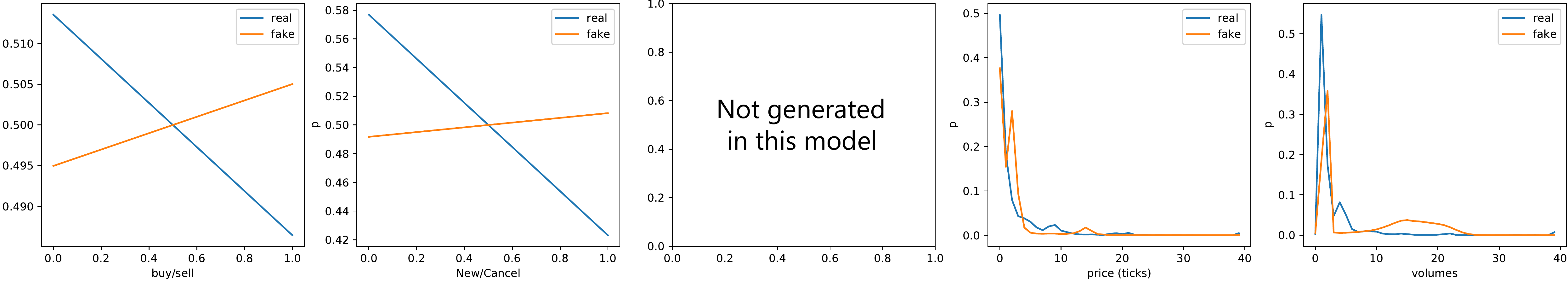}
	DCGAN 9532 JP
	\includegraphics[width=\linewidth]{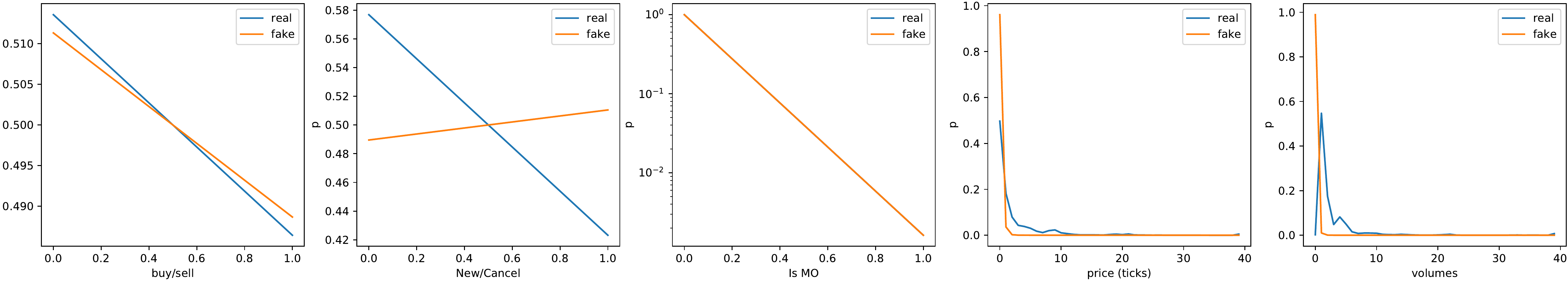}
	\caption{The results of real and fake data distribution from each models in 9532 JP.}
	\label{fig:fake-9532}
\end{figure*}

\begin{figure*}[htbp]
	\centering
	PGGAN 7012 JP\\
	\includegraphics[width=\linewidth]{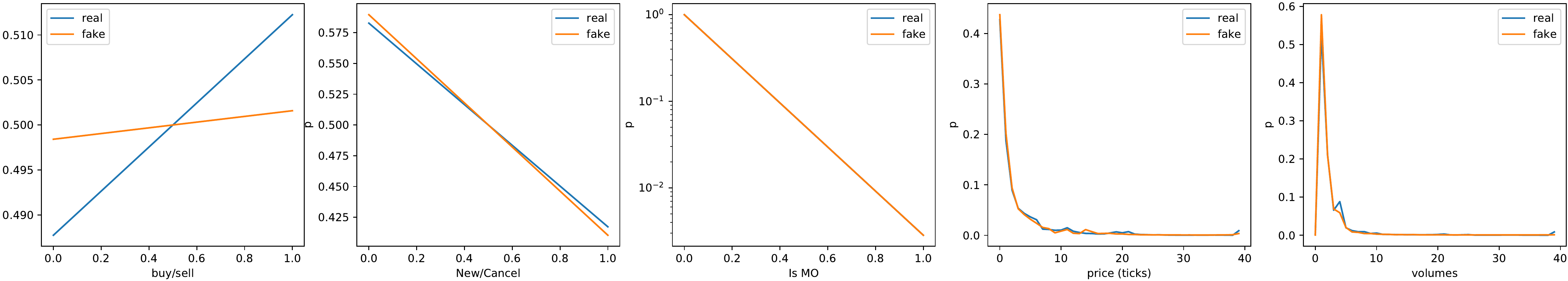}
	PGGAN-HL 7012 JP\\
	\includegraphics[width=\linewidth]{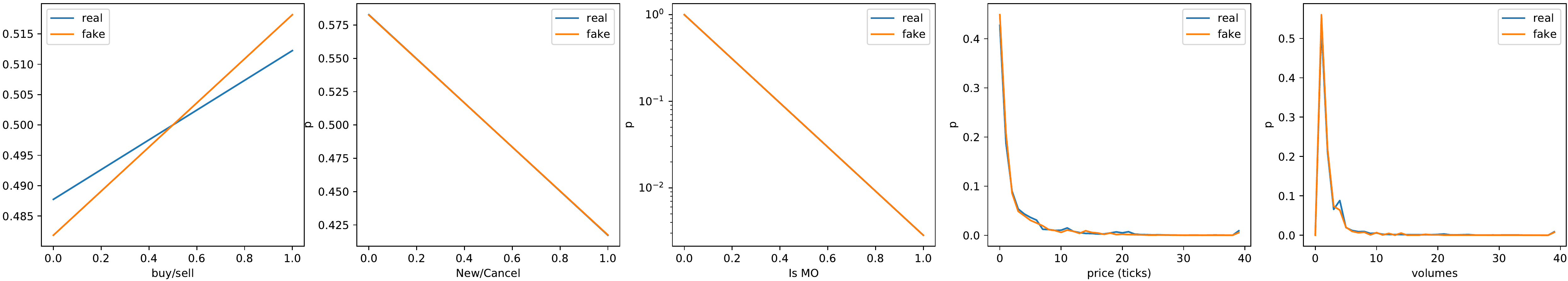}
	S-GAN 7012 JP\\
	\includegraphics[width=\linewidth]{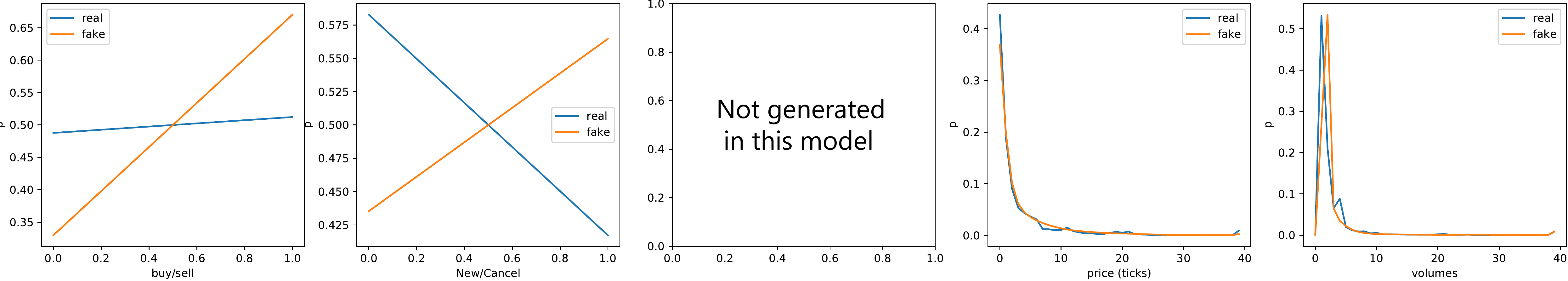}
	DCGAN 7012 JP
	\includegraphics[width=\linewidth]{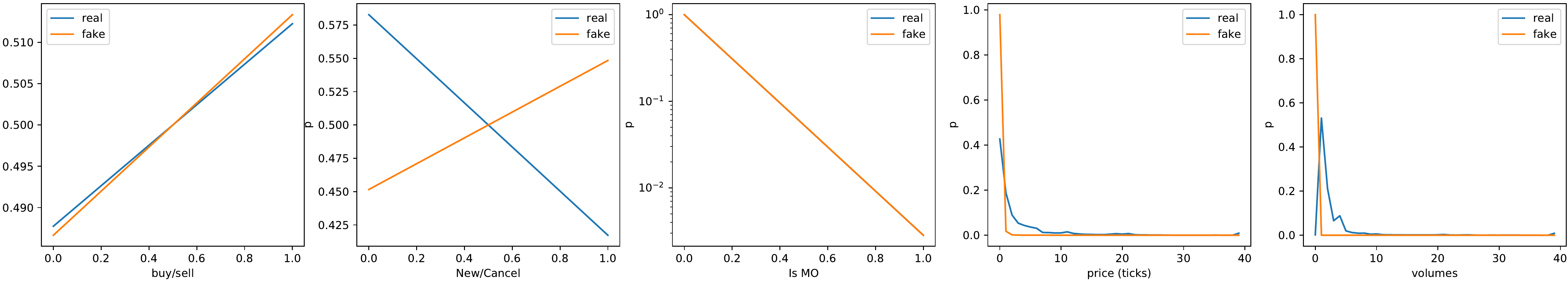}
	\caption{The results of real and fake data distribution from each models in 7012 JP.}
	\label{fig:fake-7012}
\end{figure*}

\begin{figure*}[htbp]
	\centering
	PGGAN 2501 JP\\
	\includegraphics[width=\linewidth]{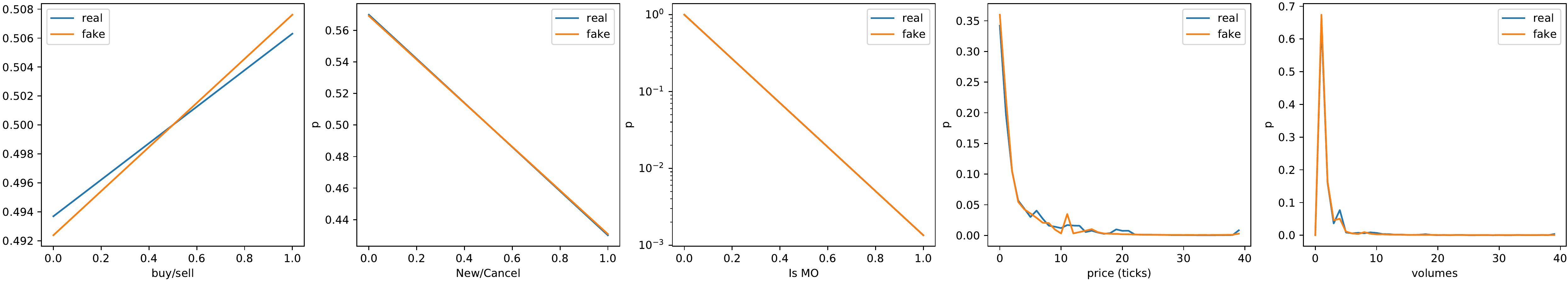}
	PGGAN-HL 2501 JP\\
	\includegraphics[width=\linewidth]{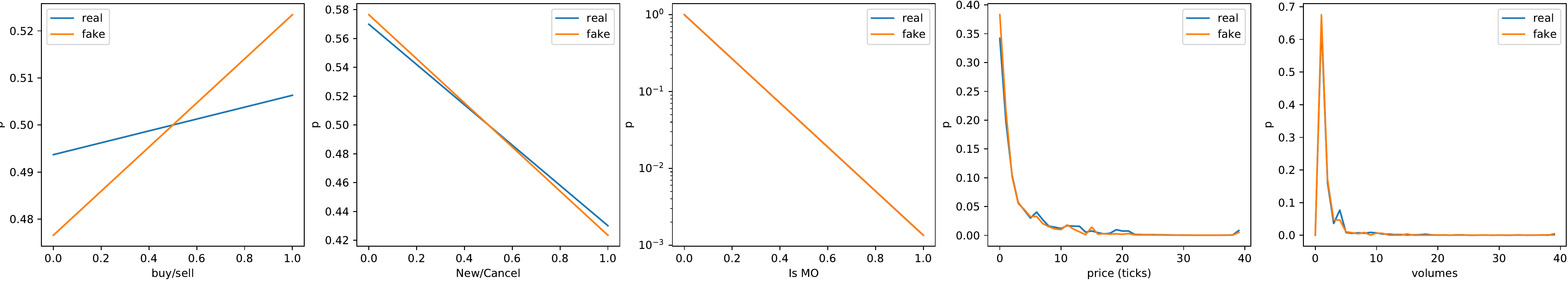}
	S-GAN 2501 JP\\
	\includegraphics[width=\linewidth]{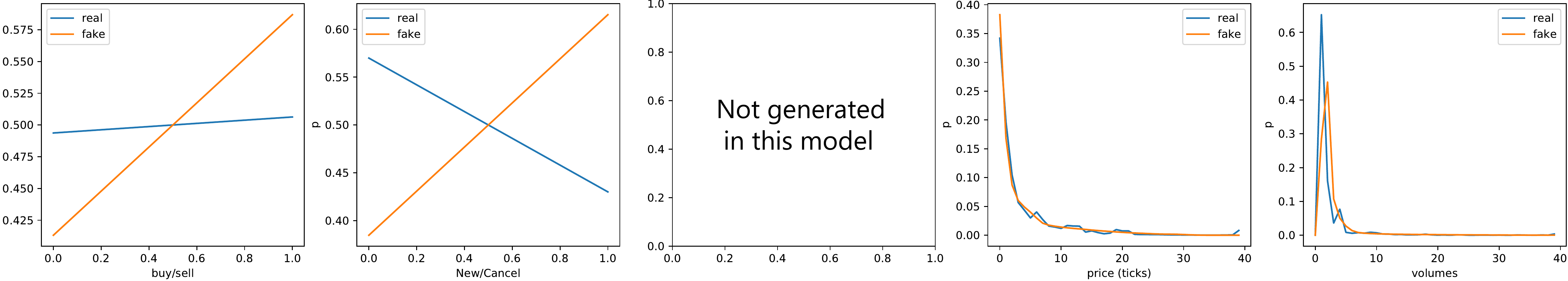}
	DCGAN 2501 JP
	\includegraphics[width=\linewidth]{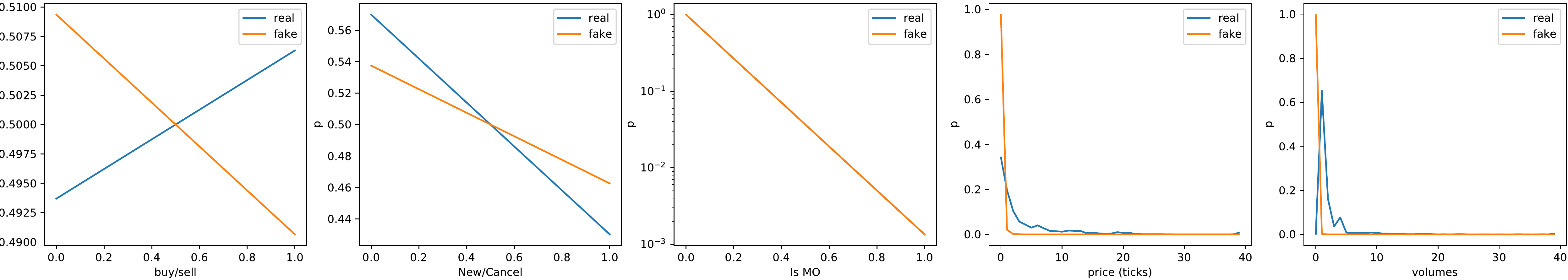}
	\caption{The results of real and fake data distribution from each models in 2501 JP.}
	\label{fig:fake-2501}
\end{figure*}

\begin{figure*}[htbp]
	\centering
	PGGAN 4005 JP\\
	\includegraphics[width=\linewidth]{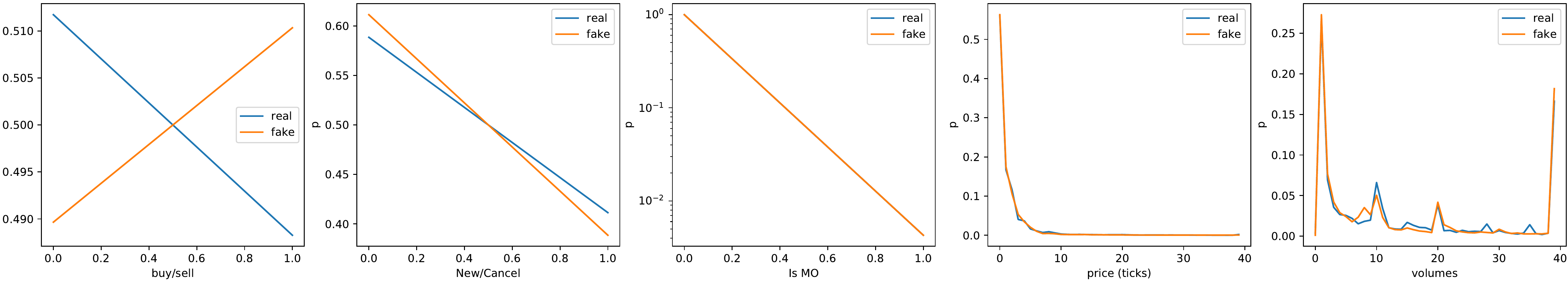}
	PGGAN-HL 4005 JP\\
	\includegraphics[width=\linewidth]{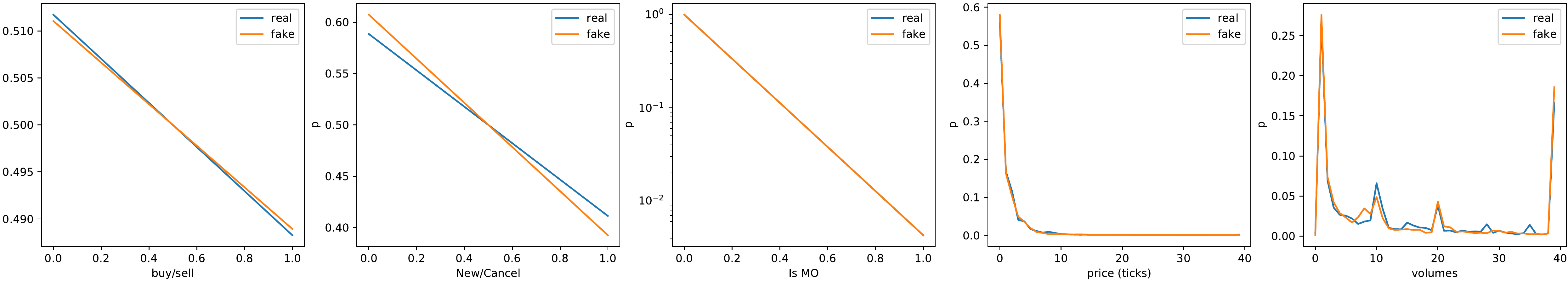}
	S-GAN 4005 JP\\
	\includegraphics[width=\linewidth]{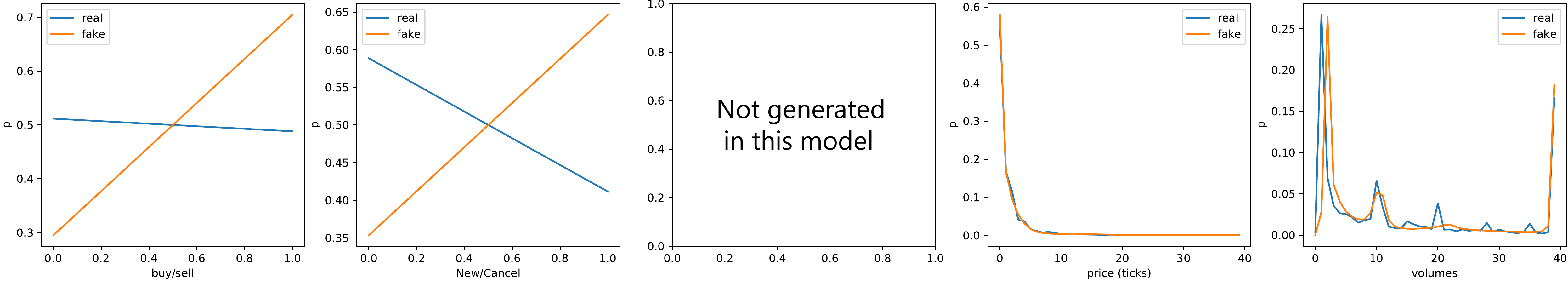}
	DCGAN 4005 JP
	\includegraphics[width=\linewidth]{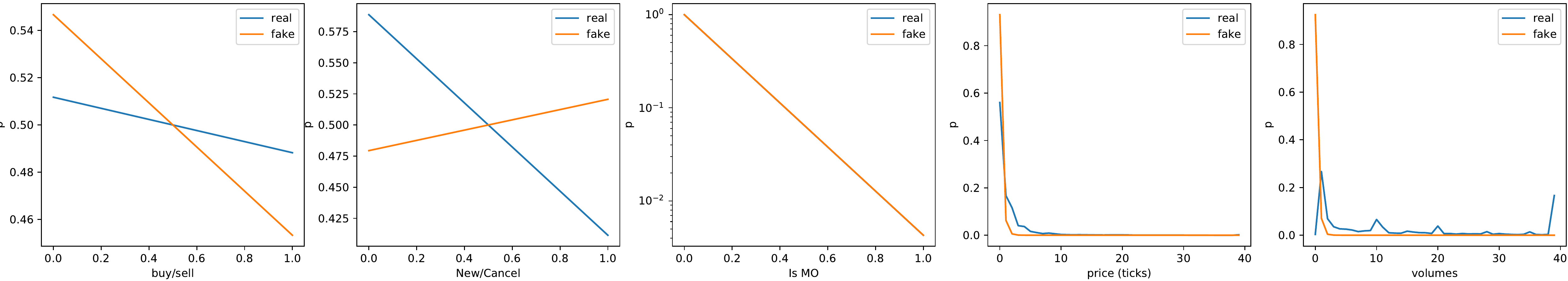}
	\caption{The results of real and fake data distribution from each models in 4005 JP.}
	\label{fig:fake-4005}
\end{figure*}

\begin{figure*}[htbp]
	\centering
	PGGAN 7752 JP\\
	\includegraphics[width=\linewidth]{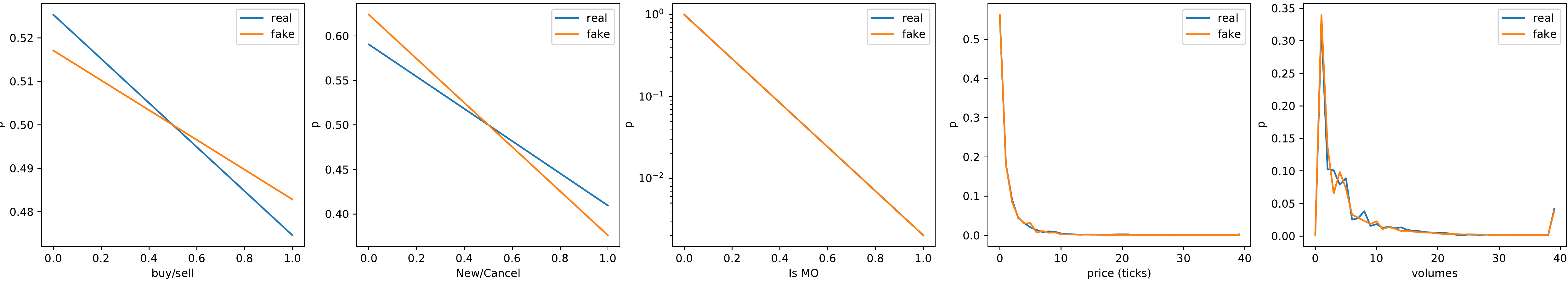}
	PGGAN-HL 7752 JP\\
	\includegraphics[width=\linewidth]{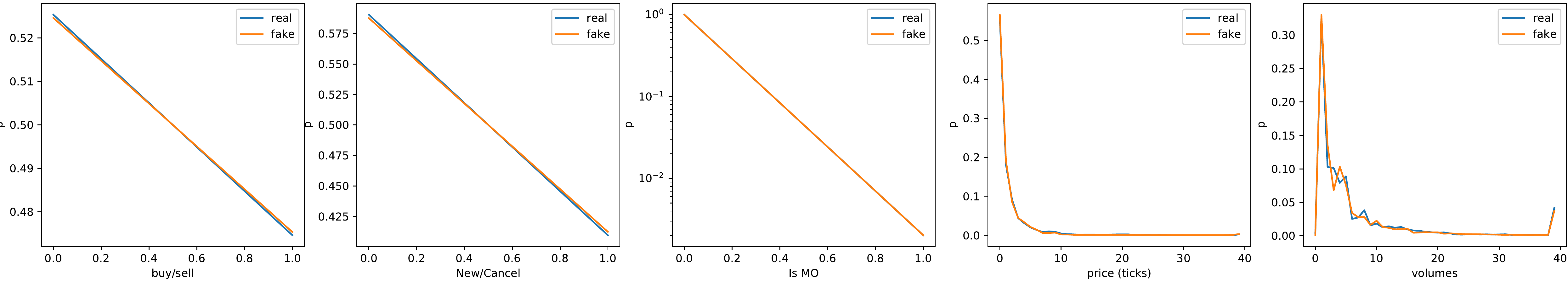}
	S-GAN 7752 JP\\
	\includegraphics[width=\linewidth]{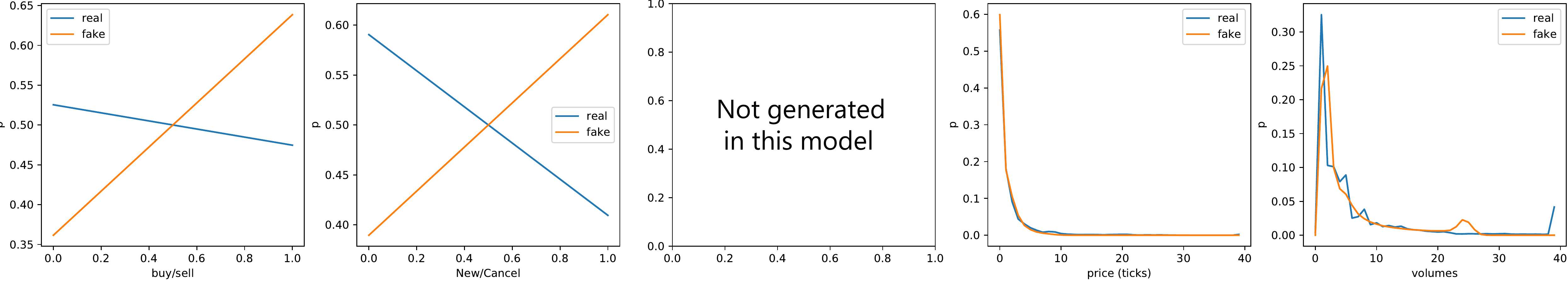}
	DCGAN 7752 JP
	\includegraphics[width=\linewidth]{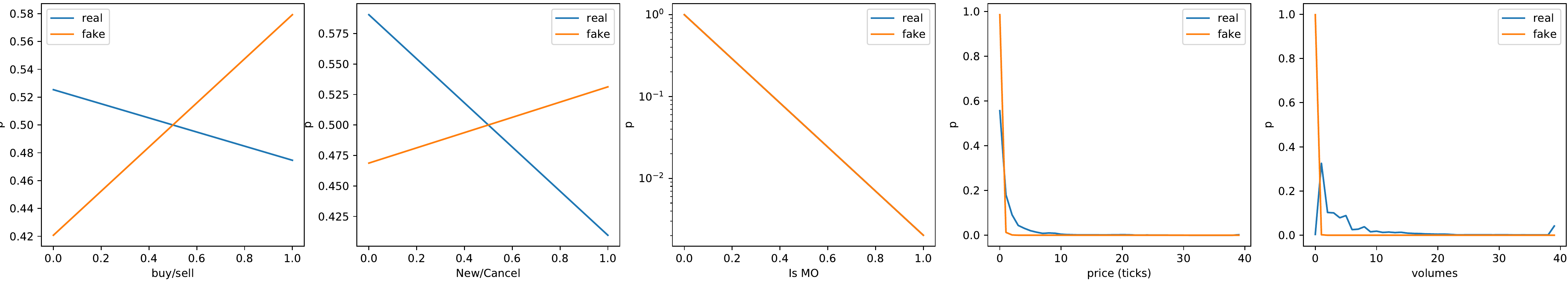}
	\caption{The results of real and fake data distribution from each models in 7752 JP.}
	\label{fig:fake-7752}
\end{figure*}

\begin{figure*}[htbp]
	\centering
	PGGAN 7911 JP\\
	\includegraphics[width=\linewidth]{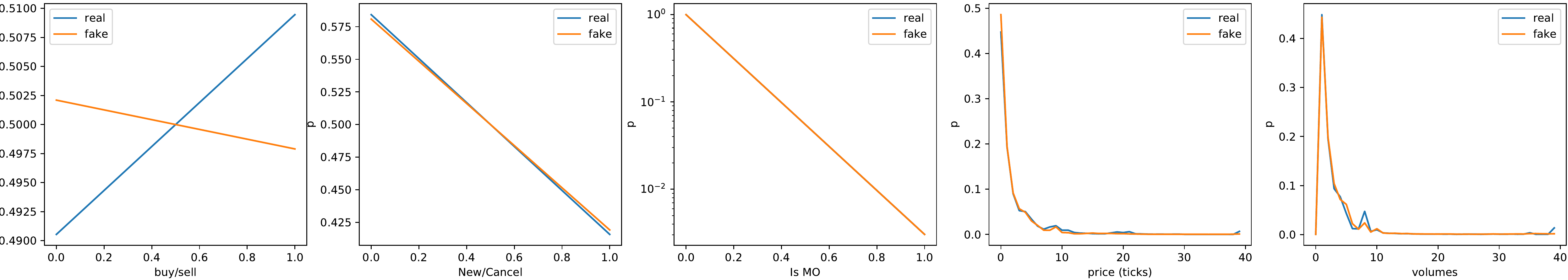}
	PGGAN-HL 7911 JP\\
	\includegraphics[width=\linewidth]{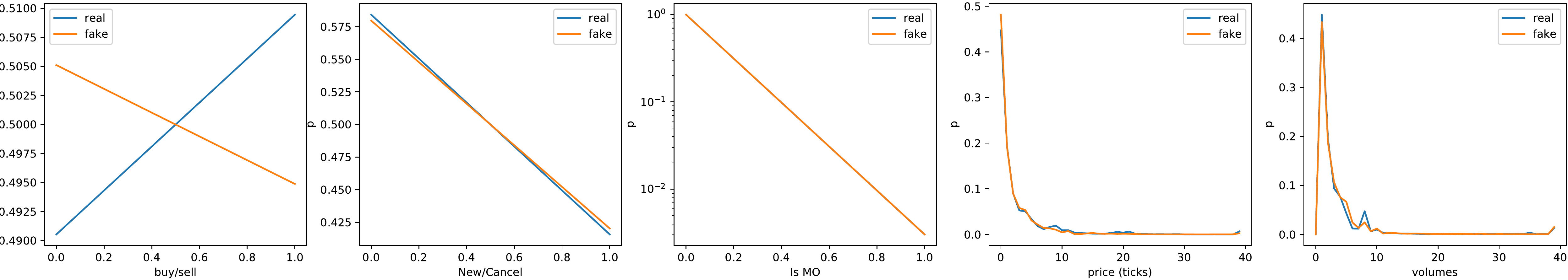}
	S-GAN 7911 JP\\
	\includegraphics[width=\linewidth]{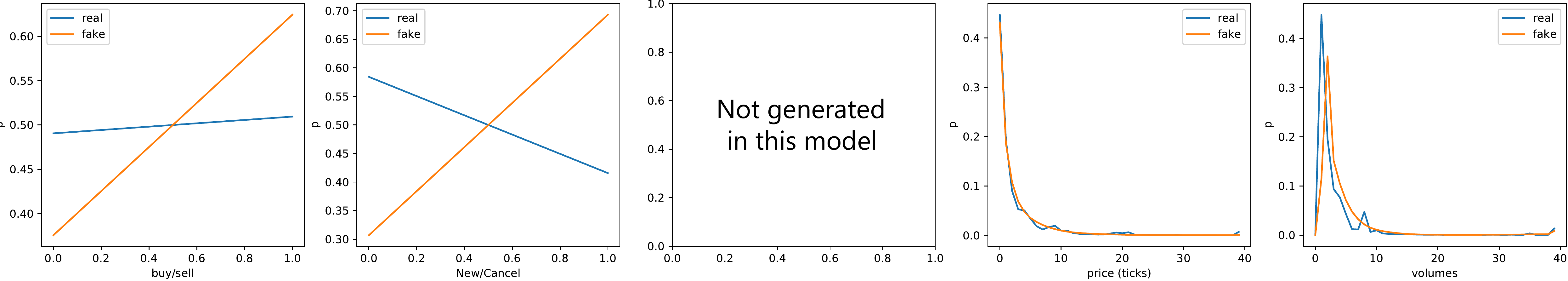}
	DCGAN 7911 JP
	\includegraphics[width=\linewidth]{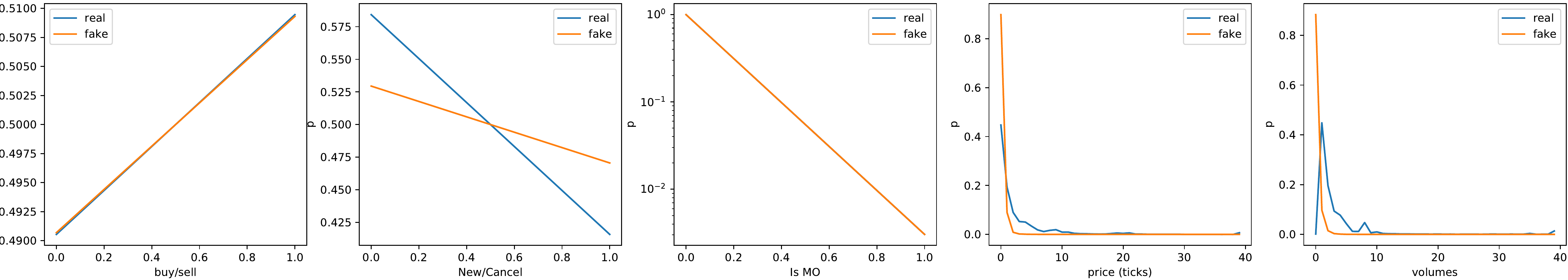}
	\caption{The results of real and fake data distribution from each models in 7911 JP.}
	\label{fig:fake-7911}
\end{figure*}

Here, we only show all distribution graphs.
Figures \ref{fig:fake-5901} -- \ref{fig:fake-7911} shows the all graphs for each ticker and models.


\end{document}